\documentclass[11pt]{article}
\usepackage[preprint]{acl}
\usepackage{times}
\usepackage{latexsym}
\usepackage[T1]{fontenc}
\usepackage[utf8]{inputenc}
\usepackage{microtype}
\usepackage{graphicx}
\usepackage{booktabs}
\usepackage{multirow}
\usepackage{tabularx}
\usepackage{array}
\usepackage{amsmath}
\usepackage{amssymb}
\usepackage{enumitem}
\usepackage{xcolor}
\usepackage{pifont}
\usepackage{listings}
\usepackage{tikz}
\usetikzlibrary{arrows.meta,calc,positioning}

\usepackage{amsmath,amsfonts,bm}

\def\eqref#1{equation~\ref{#1}}

\def\1{\bm{1}}

\DeclareMathAlphabet{\mathsfit}{\encodingdefault}{\sfdefault}{m}{sl}
\SetMathAlphabet{\mathsfit}{bold}{\encodingdefault}{\sfdefault}{bx}{n}

\definecolor{todo-red}{RGB}{185,28,28}
\definecolor{note-blue}{RGB}{29,78,216}
\newcommand{\system}{\textsc{WebAnything}}

\title{\texorpdfstring{\raisebox{-0.2em}{\includegraphics[height=1.25em,trim=252bp 136bp 209bp 141bp,clip]{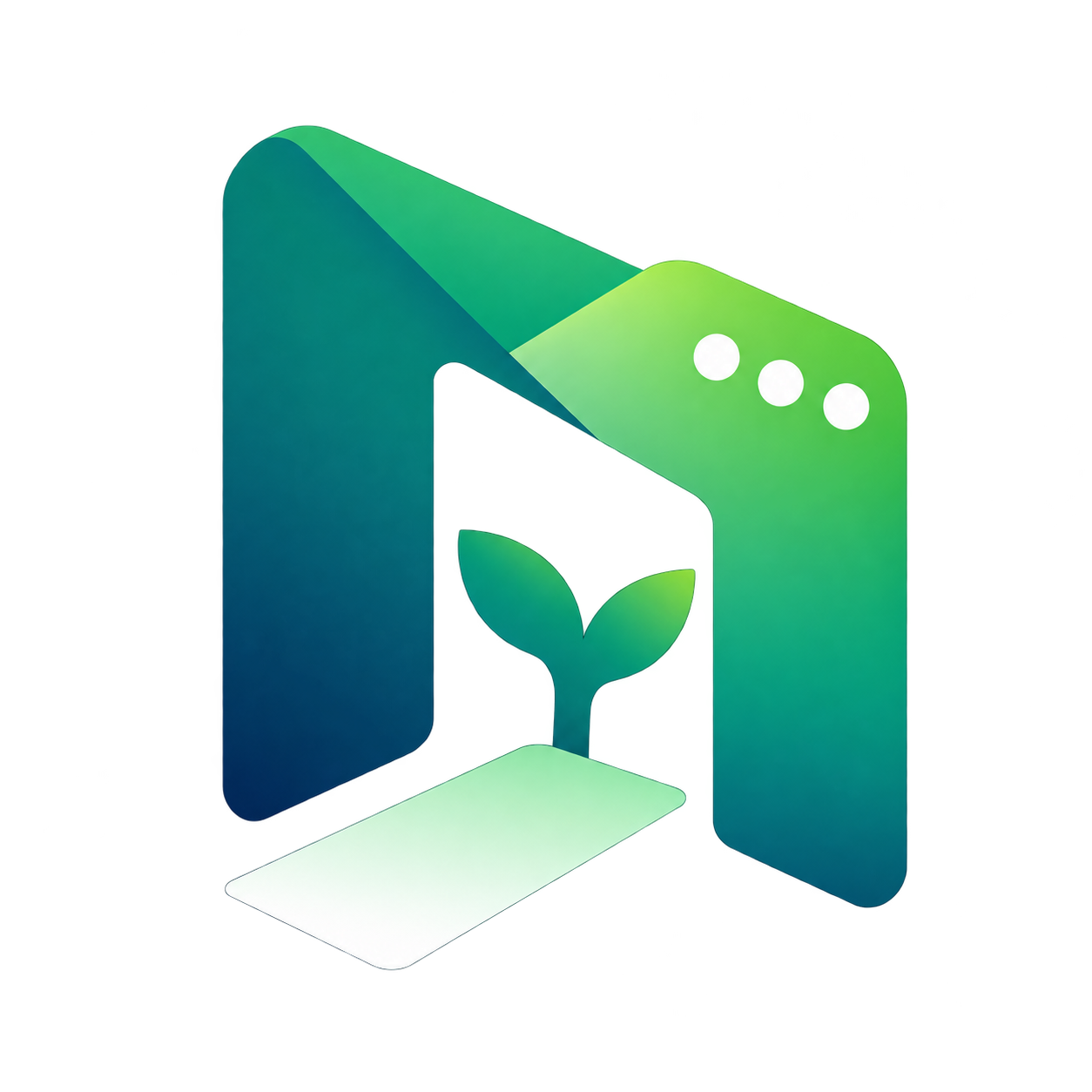}}\hspace{0.25em}}{}WebAnything: From the Wild Web to Scalable Agent Environments}
\hypersetup{
  pdftitle={WebAnything: From the Wild Web to Scalable Agent Environments},
  pdfauthor={Tianci Xie, Zikun Cai, Chaoyun Zhang, Pinxue Guo, Bingxuan Li, Rui Zheng, Hang Yan}
}
\author{
  Tianci Xie\textsuperscript{1}\quad
  Zikun Cai\textsuperscript{2}\quad
  Chaoyun Zhang\textsuperscript{3,*}\quad
  Pinxue Guo\textsuperscript{3}\\
  \bfseries Bingxuan Li\textsuperscript{3}\quad
  Rui Zheng\textsuperscript{3}\quad
  Hang Yan\textsuperscript{3}\\[0.5em]
  \normalfont\normalsize
  \textsuperscript{1}Zhejiang University\quad
  \textsuperscript{2}Nanyang Technological University\\
  \normalfont\normalsize
  \textsuperscript{3}Shanghai Qiji Zhifeng Co., Ltd\\[0.3em]
  \normalfont\small \textsuperscript{*}Corresponding author
}
\begin{document}
\maketitle

\begin{abstract}
Web agents require diverse, executable, and resettable
website environments for reproducible training and evaluation.
Human-assisted mirrors demand per-site engineering, while
generative construction can diverge from specified real websites.
Capture-and-replay preserves observations, with interaction
coverage constrained by captured content and replay capabilities.

We present \system{}, a fully automated framework that
reconstructs source-grounded, reusable Web environments
from real URLs without per-site manual implementation
or repair after initial setup and authorization.
An Explorer--Builder collects browser-observable structures,
resources, and interaction outcomes; an independent verifier
checks frozen candidates and guides local repair or
targeted source recapture.

We construct 244 website artifacts across 14 application
domains and 53 categories, supporting 12,984 executed
and accepted tasks.
Compared with human-assisted WebHarbor, \system{}
achieves higher aggregate scores on three fidelity
metrics, with 12 wins, two ties, and no losses in a
blind audit.
Fine-tuning Qwen3.5-9B on 5,856 interaction
trajectories improves REAL success by 9.9 percentage
points, Odysseys-Easy Rubric Avg and Task Avg by
6.25 and 6.32 percentage points, and WebArena-Verified
Overall success by 13.33 percentage points under the fixed harness.
Code: \href{https://github.com/china-qijizhifeng/Web-Anything}{Web-Anything on GitHub}.
\end{abstract}

\section{Introduction}
\label{sec:introduction}

Visual Web agents need stable, controllable environments in which they can explore, change state, recover from errors, and receive execution feedback. WebArena and VisualWebArena demonstrate the value of self-hosted websites for reproducible evaluation~\citep{zhou2023webarena,koh2024visualwebarena}; WebGym and WebFactory further study scaling environments and interaction experience~\citep{bai2026webgym,fan2026webfactory}. The real Web offers diverse structures and workflows, but live execution complicates state initialization, reproducibility, and safe isolation. 

We investigate how specified real websites can be
automatically transformed at scale into
source-grounded, executable, and reusable Web agent
environments, without requiring manual
site-by-site reconstruction.

\begin{figure*}[t]
\centering
\input{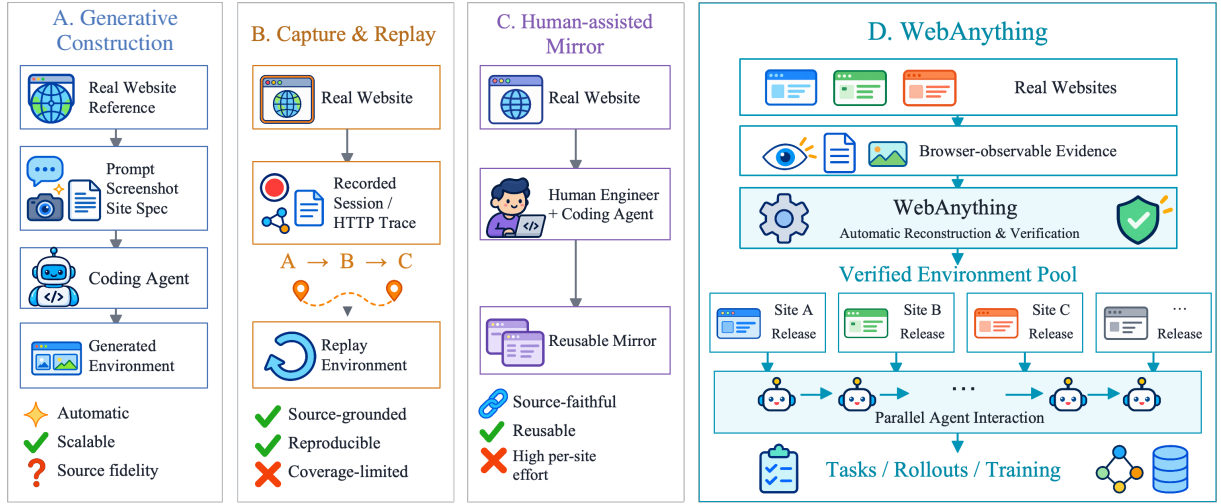}\par
\caption{Trade-offs in real-website environment construction. Human-assisted mirrors require per-site engineering; generative methods need source correspondence; replay coverage depends on captured content and replay capabilities. \system{} combines source-grounded reconstruction, independent verification, and iterative refinement.}
\label{fig:motivation}
\end{figure*}

Existing approaches trade off \textbf{automation, source fidelity, and interaction coverage} (Figure~\ref{fig:motivation}). Human-assisted mirrors provide controllable, source-corresponding environments but require per-site engineering~\citep{webharbor2026}. Generative methods scale executable sites from specifications and references~\citep{fan2026webfactory,zhang2026infiniteweb}, but matching a specified source remains a distinct requirement. Capture-and-replay preserves source observations, with interaction coverage bounded by captured content and replay capabilities~\citep{shi2017worldofbits,srivastava2025warcbench}; Weblica combines replay with source-grounded synthesis~\citep{kar2026weblica}. Scaling real-website environments requires both source correspondence and flexible, repeatable interaction.

Two technical challenges arise. The \textbf{Source-Grounding Gap} concerns reconstructing structures, resources, and behaviors when private backends and website internals are inaccessible. Screenshots and model priors alone can omit state relationships. Browser observations must instead be organized into evidence that constrains implementation. The \textbf{Verification and Feedback Gap} concerns detecting and correcting reconstruction discrepancies. Builder self-checks can overlook source-inconsistent defects, and a discrepancy may reflect either faulty implementation or insufficient observation. These cases require different actions: local repair or additional source capture.

We present \system{}, a framework for \textbf{scalable, source-grounded Web agent environment construction}. Starting from real URLs, a coupled Explorer--Builder captures page structures, resources, and before/after interaction observations, then implements executable local routes and state transitions. An Independent Verifier checks frozen candidates in a separate context against source evidence and site contracts. Evidence-supported implementation errors trigger local repair; unresolved evidence gaps trigger targeted recapture. Updated candidates are frozen and reverified before release within their declared scope. Fixed initial states, reset mechanisms, and task evaluators support repeated interaction and trajectory collection.

Our evaluation covers construction scale, environment quality, and downstream utility. The resource pool contains \textbf{244 website artifacts} across 14 domains and 53 categories. On five matched WebHarbor sites and 14 historical states, \system{} achieves higher automated fidelity metrics and 12 wins, two ties, and no losses in a blind overall-fidelity audit. Separate interaction tests characterize functional coverage. Historical analysis identifies evidence-supported discrepancies in \textbf{39 pairs across 22 sites} and independently reverified resolution of 18 issue records across six sites. The environments support \textbf{12,984 executed and accepted tasks}; SFT on earlier interaction trajectories improves Qwen3.5-9B on REAL, Odysseys-Easy, and WebArena-Verified Overall.

Our contributions are:
\begin{itemize}[leftmargin=1.3em]
\item \textbf{Fully automated, source-grounded reconstruction.}
We introduce a unified Explorer--Builder that automatically
reconstructs executable Web environments from real URLs,
using browser-observable structures, resources, and behaviors
without per-site manual implementation or debugging.

\item \textbf{Independent verification and evidence-aware refinement.}
An independent verifier checks frozen candidates against
source evidence and site contracts, directing discrepancies
to local repair or targeted recapture, followed by
automated revision and reverification.

\item \textbf{Scalable environment and task supply.}
We construct 244 website artifacts across 14 domains,
supporting 12,984 executed and accepted tasks.
Historical construction records, independent quality
evaluations, and downstream SFT demonstrate automation,
source fidelity, and practical utility.
\end{itemize}

\section{Related Work}
\label{sec:related_work}

\paragraph{Real-Web tasks and executable environments.}
Mind2Web, WebVoyager, WebGym, WebLINX, SeeAct, Online-Mind2Web, and AssistantBench study tasks or interactions on real websites~\citep{deng2023mind2web,he2024webvoyager,bai2026webgym,lu2024weblinx,zheng2024seeact,xue2025onlinemind2web,yoran2024assistantbench}. WebArena, VisualWebArena, REAL, and WebShop emphasize controlled execution and state initialization~\citep{zhou2023webarena,koh2024visualwebarena,garg2025real,yao2022webshop}. WorkArena, WorkArena++, and BrowserGym support enterprise workflows and common evaluation infrastructure, while OSWorld and AgentStudio extend executable evaluation to computer environments~\citep{drouin2024workarena,boisvert2024workarenaplus,le2025browsergym,xie2024osworld,zheng2024agentstudio}. \system{} connects real-source diversity with repeatable execution by constructing local environments for specified websites.

\paragraph{Website generation, replay, and mirroring.}
WebFactory scales executable sites from specifications and references; InfiniteWeb uses task-centric test-driven development~\citep{fan2026webfactory,zhang2026infiniteweb}. VeriEnv provides cloned environments and deterministic SDK-based task verification; WebArena-Infinity iterates generation and testing~\citep{chae2026verienv,verienvRepo2026,webarenaInfinity2026}. WebGen-Agent combines VLM appearance feedback and GUI testing with backtracking and best-candidate selection for instruction-driven website generation~\citep{lu2026webgenagent}. Our refinement grounds frozen candidates in URL-specific evidence and routes discrepancies by evidence sufficiency. Screenshot-to-code methods include pix2code, Pix2Struct, WebSight, and Design2Code~\citep{beltramelli2017pix2code,lee2023pix2struct,laurencon2024websight,si2025design2code}. World of Bits and WARC-Bench preserve captured content through replay~\citep{shi2017worldofbits,srivastava2025warcbench}; Weblica combines replay with source-grounded synthesis~\citep{kar2026weblica}. WebHarbor constructs mirrors through human--Coding Agent collaboration~\citep{webharbor2026}. \system{} jointly uses structural, resource, and behavioral evidence. Separate visual and interaction measurements complement image metrics such as SSIM and LPIPS~\citep{wang2004ssim,zhang2018lpips}.

\paragraph{Feedback and interaction-data supply.}
Self-Refine and Reflexion use feedback for output revision and agent reflection~\citep{madaan2023selfrefine,shinn2023reflexion}. Our feedback targets frozen website implementations and missing source evidence. AutoGUIWorld synthesizes GUI observations and trajectories, while WEFT studies coordination among environments, tasks, harnesses, and evaluators~\citep{yang2026autoguiworld,mao2026weft}. Synatra and AgentTrek derive demonstrations from indirect knowledge or tutorials; WebRL studies self-evolving task curricula~\citep{ou2024synatra,xu2024agenttrek,qi2024webrl}. OS-Atlas, UI-TARS, and OpenCUA expand GUI grounding or interaction data~\citep{wu2024osatlas,qin2025uitars,wang2025opencua}. \system{} supplies executable environments and collects interaction data with the fixed agent harness and a unified task/evaluator pipeline, measuring utility on external benchmarks.

\section{Methodology}
\label{sec:method}

\begin{figure*}[t]
\centering
\includegraphics[width=\textwidth]{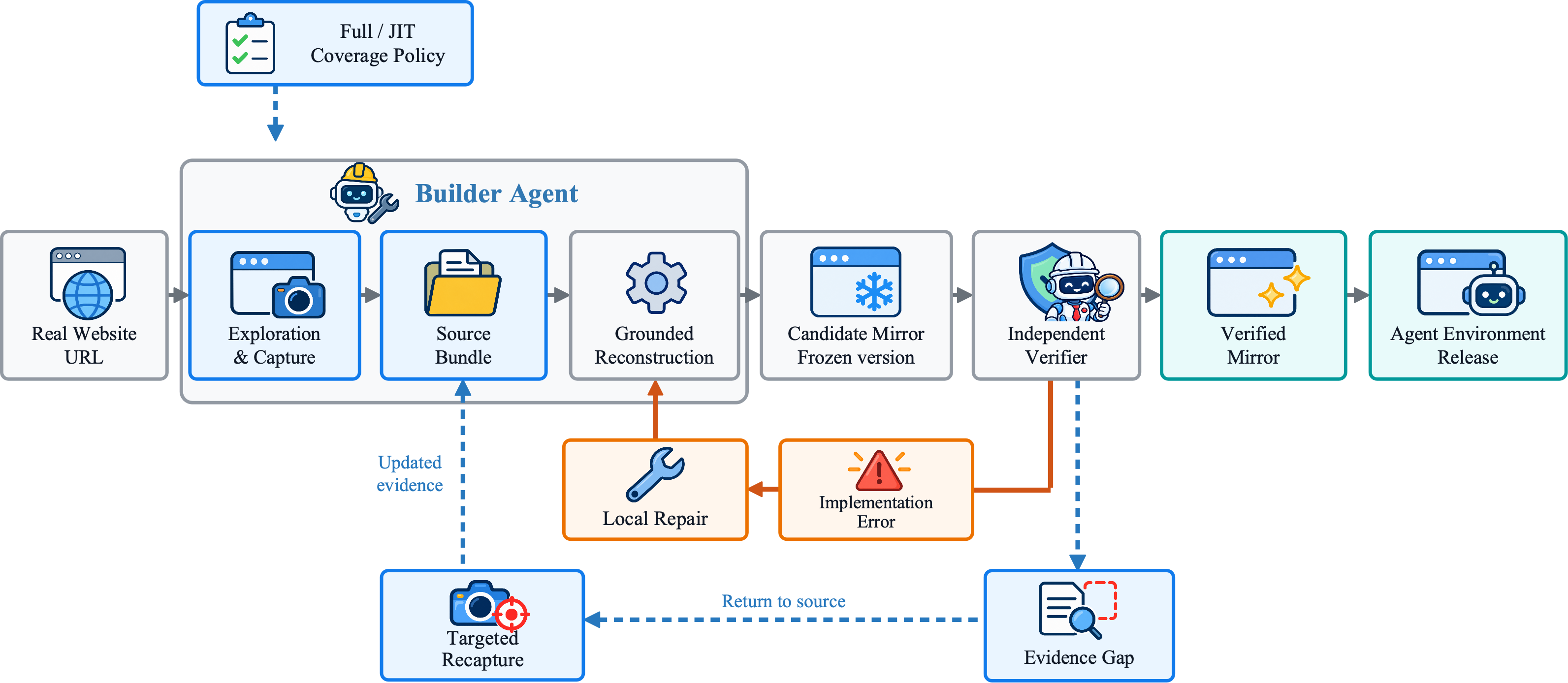}
\caption{\system{} reconstruction pipeline. Source evidence guides executable construction. Independent checks of frozen candidates drive local repair or targeted recapture; verified candidates are released with repeatable interaction and task-evaluation interfaces.}
\label{fig:system-overview}
\end{figure*}

\system{} uses source evidence as the shared basis for construction and independent verification (Figure~\ref{fig:system-overview}). An Explorer--Builder reconstructs executable local pages and states; verification feedback then drives implementation repair or targeted source recapture.

\subsection{Problem Formulation}
\label{sec:problem-formulation}

Given a website $W$, entry URL, permitted access scope, and construction budget, we seek a local environment $\widehat{W}$ preserving observed presentation, interactions, and state changes within a site contract $\mathcal{C}$. The source is accessible through browser observations rather than private backend code. Its evidence is
\begin{equation}
\mathcal{E}=(\mathcal{E}_p,\mathcal{E}_r,\mathcal{E}_b),\quad
\mathcal{E}_b=\{(o_t,a_t,o_{t+1})\}_{t=1}^{N},
\end{equation}
where $\mathcal{E}_p$ records page structure and appearance, $\mathcal{E}_r$ resources and dependencies, and $\mathcal{E}_b$ observations before and after actions. The contract specifies required pages, workflows, and checkable outcomes. Reconstruction translates these observations into executable local state transitions that support different tasks within the implemented scope.

\subsection{Source-Grounded Reconstruction}
\label{sec:source-reconstruction}

\paragraph{State-associated evidence.}
For each visited state, the Explorer--Builder associates page structures, computed styles, element geometry, screenshots, and resource references. Actions are linked to subsequent visible feedback and state changes. Distinct interaction states at the same URL remain distinct evidence: a closed menu and its expanded state, for example, constrain different implementation requirements.

\paragraph{Evidence-constrained implementation.}
Structural observations constrain layout and component relationships; resource observations ground fonts, images, and media; behavioral observations constrain navigation, filtering, and state modification. The Builder organizes these into local components, routes, and shared object relationships. Screenshot references thus complement structural and behavioral evidence. A specific discrepancy can be traced back to its associated page state and action. When available observations do not establish a required outcome, the system seeks additional evidence rather than treating an inferred behavior as observed.

\paragraph{Coverage policies.}
\textbf{Full Coverage} expands accessible page types, interaction states, and neighboring branches from the entry URL and native navigation. \textbf{Just-in-Time Coverage} prioritizes dependencies of a given task set and expands them through construction and verification feedback. Both share the reconstruction and acceptance mechanisms, differing in exploration scope and expansion timing. Reported construction experiments use Full Coverage.

\subsection{Independent Verification and Evidence-Aware Refinement}
\label{sec:verification-feedback}

\paragraph{Verification decoupling.}
The Builder implements and self-checks a candidate; an Independent Verifier then examines the frozen implementation in a separate context. It obtains rendering and operation results anew and compares them with source evidence and the site contract. Applicable checks cover pages, controls, resources, workflows, persistence, and restoration. The Verifier cannot edit candidate code. Stateful tests operate through local UI/API interfaces and restore the initial state. Candidate freezing binds findings to the inspected implementation and prevents verification-time repair from changing the target.

\paragraph{Evidence-aware refinement.}
The model diagnoses each discrepancy using check results, source observations, the contract, and runtime conditions. If existing evidence establishes an expected outcome that the candidate fails to reproduce, \textbf{Local Repair} returns the affected page or control, executed action, expected and observed outcomes, and supporting evidence to the Builder. If evidence is insufficient to establish the target outcome, \textbf{Targeted Recapture} forms a concrete observation request and revisits the relevant source region. Newly captured structures, resources, or behaviors constrain the next implementation.

For example, Figma's authorized source capture established a canvas-anchored comment composer. Independent checks found mismatched placement and appearance, providing a concrete target for local revision. The Builder read the feedback, changed the editor markup and styles, and submitted a new candidate; independent checks then confirmed the composer's geometry and visibility. Existing source evidence supported repair in this case; an unresolved observation question instead becomes a bounded recapture request (Appendix~\ref{sec:verification-case-evidence}).

After either path updates the candidate, the controller freezes it again and the Verifier repeats affected checks and applicable regression tests. Verification therefore connects discrepancies to actionable revision, while source evidence constrains both initial reconstruction and subsequent refinement. Candidates passing checks within their declared scope can proceed to release. Implementation details and original prompts appear in Appendix~\ref{sec:verifier-implementation-audit}.

\subsection{Environment and Verifiable Task Supply}
\label{sec:environment-task-supply}

Released environments package local implementations, resources, initial data, runtime configuration, and reset interfaces. Shared state connects related pages, and applicable creation, modification, and save operations affect later interactions. Fixed initial data and state restoration provide repeatable task starting conditions. The task pipeline generates instructions and matching success conditions for implemented capabilities. Agents execute tasks through a fixed browser-based computer-use agent harness shared across data collection and downstream evaluations; settings and execution budgets are specified in Appendix~\ref{sec:transfer-protocol}. Evaluators check final answers or terminal environment states. Accepted trajectories are converted into step-level training examples. Task generation, executed acceptance, trajectory collection, and training export remain separate stages and counting units; Section~\ref{sec:downstream-utility} reports their outputs.

\section{Experiments and Results}
\label{sec:experiments}

We study \textbf{RQ1: construction scale and efficiency}, \textbf{RQ2: environment quality and independent verification}, and \textbf{RQ3: task support and downstream learning utility}.

\subsection{Experimental Setup}

\paragraph{Objects and baselines.}
The resource inventory covers 244 unique website artifacts as of October 9, 2026. Full Coverage construction records for 85 sites support automation and expenditure analysis. Quality comparisons use frozen artifacts and an independent evaluation pipeline: five WebHarbor sites with 14 historical reference states, and five VeriEnv artifacts for offline interaction tests. Historical VeriEnv fidelity references cover four LinkedIn states. Subsets and reference conditions are specified in Appendices~\ref{sec:artifact-eligibility}--\ref{sec:current-source-spotcheck}.

\paragraph{Measurements.}
Pixel Similarity, SSIM, and Text F1 measure source-state correspondence, aggregated within sites and then averaged equally across sites. A blind audit of the 14 WebHarbor-matched states uses final labels agreed through discussion by three annotators, with randomized anonymous mirror order. Interaction tests disable external network access and report capability coverage, executed-test success, and all-test pass, keeping their denominators distinct. Explicit adapters restore initial states. Historical verification analysis links self-checks, independent findings, and repair traces; its definitions appear in Appendix~\ref{sec:verification-analysis-details}.

\paragraph{Tasks and training.}
Two disjoint collections contain 6,172 accepted tasks across 74 sites and 6,812 across 229 sites. SFT uses earlier WebAnything interaction data: 5,856 trajectories associated with 5,167 unique training task IDs, converted into 60,517 step-level examples. Multiple trajectories or format variants can share a task ID. The newer 6,812 tasks are not used in this SFT experiment. We fine-tune Qwen3.5-9B for 473 steps (the fixed training endpoint), with global batch size 256 and learning rate $5\times10^{-6}$. Evaluation uses this trained model on REAL~\citep{garg2025real}, Odysseys-Easy~\citep{jang2026odysseys}, and WebArena-Verified~\citep{elhattami2025webarena}; configurations and overlap checks appear in Appendix~\ref{sec:transfer-protocol}.

\subsection{RQ1: Construction Scale and Efficiency}
\label{sec:scalable-supply}

\begin{figure*}[t]
\centering
\includegraphics[width=\textwidth]{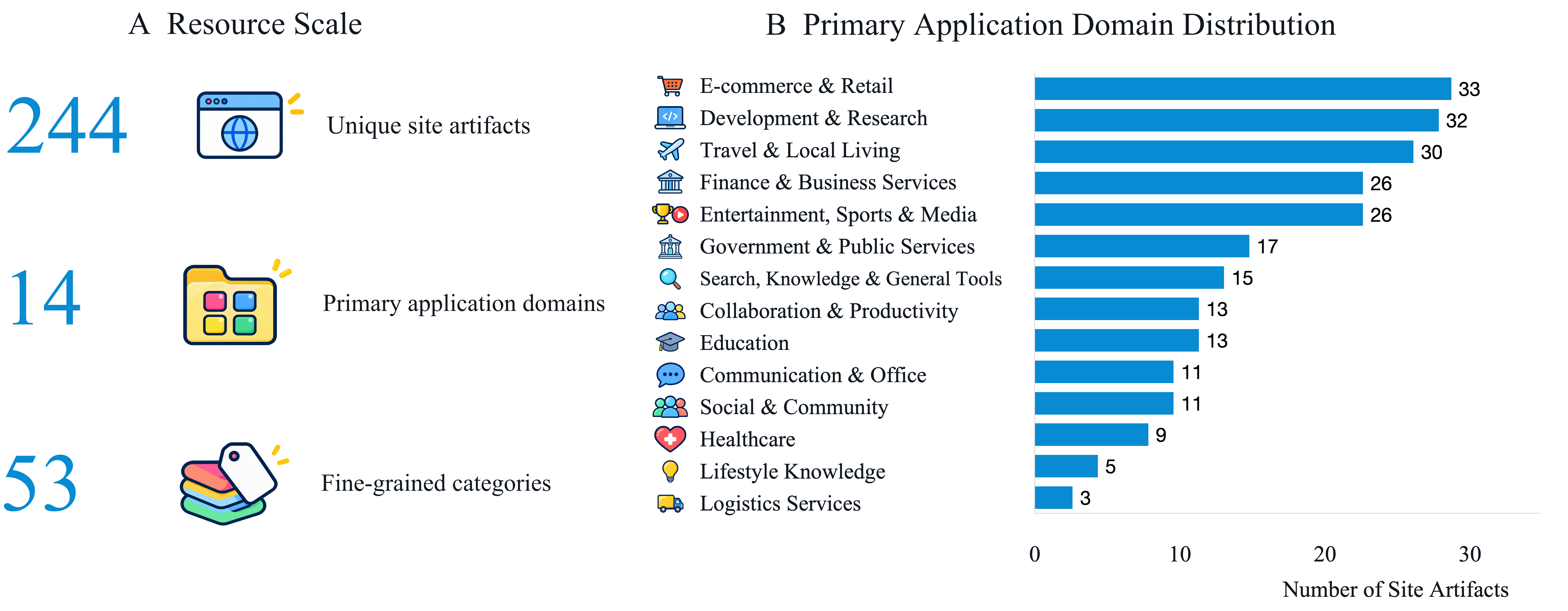}
\caption{Cross-domain resource supply: 244 website artifacts across 14 application domains and 53 fine-grained categories, with the domain distribution at right.}
\label{fig:environment-coverage}
\end{figure*}

\paragraph{Cross-domain supply.}
The unified pipeline produces \textbf{244 website artifacts} across \textbf{14 application domains and 53 categories} (Figure~\ref{fig:environment-coverage}), spanning diverse layouts, navigation structures, and workflows. Artifacts are counted by site identity and recorded homepage availability; deeper fidelity and functionality are assessed separately on evaluated subsets. Appendix~\ref{sec:environment-site-list} lists the sites, and Appendix~\ref{sec:pool-quality-depth} reports additional rendering diagnostics.

\paragraph{Automation and recorded expenditure.}
After initial configuration and necessary authorization, the 85-site historical collection contains no recorded per-site manual implementation, debugging, or repair. Exploration, capture, implementation, and revisions are performed by the pipeline. Recorded model-call cost for 84 sites totals \textbf{\$220.62} (mean \textbf{\$2.63}, median \textbf{\$1.60} per site), a lower bound that excludes unpriced calls and system overhead. Historical outcomes and accounting details are in Appendix~\ref{sec:historical-construction-records}.

\subsection{RQ2: Environment Quality and Independent Verification}
\label{sec:environment-quality}

\paragraph{A. Source-state fidelity.}
On five matched WebHarbor sites and 14 historical states, \system{} scores higher on all three aggregate metrics: Pixel Similarity \textbf{0.9240 vs.\ 0.8083}, SSIM \textbf{0.7904 vs.\ 0.6260}, and Text F1 \textbf{0.8541 vs.\ 0.2617} (Table~\ref{tab:source-fidelity}). The blind overall-fidelity audit gives \textbf{12 wins, two ties, and no losses}; separate information-fidelity and discrepancy-severity judgments show complementary differences. Human judgments concern visible reference states, while functionality is tested through execution. The four-state LinkedIn comparison also favors \system{} in Pixel and SSIM, reported separately.

Historical references include rendering defects, and source/mirror JavaScript conditions differ for Text F1. Appendix~\ref{sec:historical-fidelity-details} specifies these conditions and sensitivity results; Appendix~\ref{sec:human-state-audit} gives the blind-audit protocol. Selected page comparisons appear in Figure~\ref{fig:historical-qualitative} in the appendix.

\begin{table*}[t]
\centering
\begingroup
\renewcommand{\thetable}{\arabic{table}a}
\renewcommand{\theHtable}{historical-state-fidelity}

\small
\setlength{\tabcolsep}{4pt}
\begin{tabularx}{\linewidth}{p{0.28\linewidth}p{0.20\linewidth}*{4}{>{\centering\arraybackslash}X}}
\toprule
\multicolumn{6}{l}{\textbf{A: Automated fidelity metrics}} \\
\midrule
Comparison subset & System & Sites/states & Pixel $\uparrow$ & SSIM $\uparrow$ & Text F1 $\uparrow$ \\
\midrule
Five shared WebHarbor sites & \system{} & 5 / 14 & 0.9240 & 0.7904 & 0.8541 \\
Same subset & WebHarbor & 5 / 14 & 0.8083 & 0.6260 & 0.2617 \\
\midrule
VeriEnv LinkedIn & \system{} & 1 / 4 & 0.9574 & 0.8190 & N/A \\
Same subset & VeriEnv & 1 / 4 & 0.9436 & 0.7637 & N/A \\
\bottomrule
\end{tabularx}
\par\vspace{5pt}
\begin{tabularx}{\linewidth}{p{0.32\linewidth}*{4}{>{\centering\arraybackslash}X}}
\toprule
\multicolumn{5}{l}{\textbf{B: Human blind audit}} \\
\midrule
Assessment & Wins & Ties & Losses & Win rate $\uparrow$ \\
\midrule
Overall fidelity (Q1) & 12 & 2 & 0 & 85.7\% \\
Information fidelity (Q2) & 10 & 3 & 1 & 71.4\% \\
\midrule
System (Q3) & Equivalent & Minor & Severe & \shortstack{Mean severity\\$\downarrow$} \\
\midrule
\system{} & 12/14 & 2/14 & 0/14 & 0.143 \\
WebHarbor & 0/14 & 8/14 & 6/14 & 1.429 \\
\bottomrule
\end{tabularx}
\par\vspace{3pt}
\begin{minipage}{\linewidth}\footnotesize
Q3 codes visible discrepancies in information or interaction cues as 0 (equivalent), 1 (minor), or 2 (severe). Ratings concern historical page states.
\end{minipage}
\caption{Historical source-state fidelity. Automated metrics are aggregated within sites, then equally averaged across the five-site, 14-state WebHarbor subset; the four historical LinkedIn states are reported separately. Human blind-audit wins, ties, and losses compare \system{} against WebHarbor.}
\label{tab:source-fidelity}
\endgroup
\end{table*}

\paragraph{B. Interaction validity.}
Offline tests cover navigation, retrieval, forms, state mutations, and applicable recovery operations (Table~\ref{tab:environment-validity}). On the WebHarbor subset, \system{} achieves 75\% capability coverage and 63.3\% all-test pass, versus WebHarbor's 80\% and 76.7\%, showing broader tested interaction coverage for WebHarbor. On the separate VeriEnv subset, selected pool artifacts pass \textbf{29/33 tests}, versus \textbf{21/33} for VeriEnv; results vary by site, with VeriEnv ahead on LinkedIn. Artifact inclusion and per-site results appear in Appendix~\ref{sec:validity-probe-details}.

Together, the comparisons show that \emph{source fidelity determines what agents observe, while interaction validity determines what they can execute}. Higher visual correspondence alone does not establish broader functional coverage.

\begin{table*}[t]
\centering
\addtocounter{table}{-1}
\begingroup
\renewcommand{\thetable}{\arabic{table}b}
\renewcommand{\theHtable}{artifact-validity}
\small
\setlength{\tabcolsep}{4pt}
\begin{tabularx}{\linewidth}{Xccc}
\toprule
System & Capability coverage & Executed-test success & All-test pass \\
\midrule
\multicolumn{4}{l}{\textit{Five-site WebHarbor comparison}} \\
\system{} & 15/20 (75\%) & 19/20 (95\%) & 19/30 (63.3\%) \\
WebHarbor & 16/20 (80\%) & 23/23 (100\%) & 23/30 (76.7\%) \\
\midrule
\multicolumn{4}{l}{\textit{Five-site VeriEnv comparison}} \\
Selected pool artifacts & 22/22 (100\%) & 29/31 (93.5\%) & 29/33 (87.9\%) \\
VeriEnv & 20/22 (90.9\%) & 21/28 (75.0\%) & 21/33 (63.6\%) \\
\bottomrule
\end{tabularx}
\caption{Offline interaction validity on separate five-site subsets. Capability coverage counts tested capabilities; executed-test success uses executed tests; all-test pass retains unsupported and blocked tests. Persistence and state-restoration results appear in Appendix~\ref{sec:validity-probe-details}.}
\label{tab:environment-validity}
\endgroup
\end{table*}

\paragraph{C. Independent verification and feedback.}
\label{sec:verification-decoupling-analysis}
Of 229 historical Builder--Verifier report pairs across 69 sites, 219 have passing Builder self-checks (Appendix Table~\ref{tab:independent-verification}). Independent verification identified evidence-supported discrepancies in \textbf{39 historical report pairs across 22 sites}, including one strictly candidate-matched comparison. Findings span interactions, appearance, state, resources, and workflows. In Stashmedia, passing resource self-checks contrast with independently observed external images and broken assets in the same candidate.

A separate repair-trace cohort contains feedback, revision, and subsequent verification records for 46 sites. Issue-level inspection confirms \textbf{18 independently reverified resolution records across six sites}: 14 local implementation repairs, three runtime repairs, and one request-lifecycle issue. Ctrip illustrates five resolved conditions, including login rejection, destination search, favorites, booking drafts, and reset (Figure~\ref{fig:verification-feedback-cases} in the appendix).

These historical records document additional defect findings and traceable, independently reverified repairs. The analysis is observational; pairing, adjudication, and counting details are provided in Appendix~\ref{sec:verification-analysis-details}.

\subsection{RQ3: Task Support and Downstream Learning Utility}
\label{sec:downstream-utility}

\begin{table*}[t]
\centering

\small
\setlength{\tabcolsep}{3pt}
\renewcommand{\arraystretch}{1.15}
\begin{tabularx}{\linewidth}{p{0.25\linewidth}*{4}{>{\centering\arraybackslash}X}}
\toprule
Training setting & REAL & \multicolumn{3}{c}{Odysseys-Easy} \\
\cmidrule(lr){3-5}
 & Success $\uparrow$ & Rubric Avg $\uparrow$ & Task Avg $\uparrow$ & Perfect $\uparrow$ \\
\midrule
Qwen3.5-9B & 24.2 & 19.71 & 18.86 & 4.55 \\
+ WebAnything SFT & 34.1 & 25.96 & 25.18 & 9.09 \\
$\Delta$ & +9.9pp & +6.25pp & +6.32pp & +4.55pp \\
\bottomrule
\end{tabularx}
\par\vspace{3pt}
\begin{tabularx}{\linewidth}{p{0.25\linewidth}*{3}{>{\centering\arraybackslash}X}}
\toprule
\multicolumn{4}{l}{\textbf{WebArena-Verified: fixed agent harness}} \\
\midrule
Task type & Qwen3.5-9B & WebAnything SFT & $\Delta$ \\
\midrule
Retrieve & 20.75 & 12.00 & $-8.75$pp \\
Navigate & 22.22 & 40.71 & +18.49pp \\
Mutate & 10.51 & 42.01 & +31.50pp \\
Overall & 16.41 & 29.74 & +13.33pp \\
\bottomrule
\end{tabularx}
\par\vspace{3pt}
\begin{minipage}{\linewidth}\footnotesize
$\Delta$ denotes percentage points. Perfect counts are 2/44 and 4/44. \textbf{Same-model Retrieve rescoring:} ignoring JSON-format judgment increases SFT passes from 39/325 (12.00\%, strict) to 105/325 (32.31\%), or 66 additional passes (+20.31pp). This supplement changes scoring, not the model or agent execution.
\end{minipage}
\caption{Downstream SFT results. REAL and Odysseys-Easy use 91 and 44 jointly valid tasks. WebArena-Verified uses our own implementation of the fixed harness with strict scoring. Values are percentages.}
\label{tab:post-training-utility}
\end{table*}

\paragraph{A. Executed task supply.}
\label{sec:task-scaling}
Agents and evaluators yield \textbf{12,984 executed and accepted tasks} from two disjoint collections. The newer batch supplies \textbf{6,812 tasks across 229 sites}: 217 reach 30 tasks, 12 partially complete the quota, and 15 of the 244 pool sites have none in this batch. Completion relative to the $244\times30=7{,}320$ target is \textbf{93.1\%}. The earlier collection supplies 6,172 tasks across 74 sites, whose sites are a subset of the newer batch's sites. Acceptance definitions and snapshot statistics are in Appendix~\ref{sec:task-supply-snapshot-protocol}.

\paragraph{B. Diversity and complexity.}
The earlier tasks cover retrieval, search/filtering, comparison, and state operations. Automated, overlapping capability labels mark 26.8\% for state operations and 37.7\% for cross-site attributes. Observed tool-interaction depth averages \textbf{10.4 steps} (median 8, 90th percentile 21, maximum 89); \textbf{34.7\%} exceed 10 steps. These executions demonstrate support for queries and multi-step workflows. Figure~\ref{fig:task-diversity-complexity} and Appendix~\ref{sec:verified-task-diversity} provide full distributions and label definitions.

\paragraph{C. Downstream learning.}
SFT on the 5,856 WebAnything interaction trajectories improves performance on three external benchmarks (Table~\ref{tab:post-training-utility}). On 91 jointly valid REAL tasks, success increases from \textbf{22/91 to 31/91} (+9.9 percentage points). Retaining infrastructure failures in the full 102-task set gives 21.6\% vs.\ 30.4\% (+8.8pp). On 44 jointly valid Odysseys-Easy tasks, Rubric Avg rises from \textbf{19.71\% to 25.96\%} (+6.25pp), Task Avg from \textbf{18.86\% to 25.18\%} (+6.32pp), and fully successful tasks from 2/44 to 4/44.

On WebArena-Verified, the fixed harness reports an Overall gain of \textbf{13.33pp} (16.41\% to 29.74\%), with substantial gains on Navigate (+18.49pp) and Mutate (+31.50pp). Retrieve decreases by 8.75pp under strict scoring. Supplementary rescoring of the same SFT outputs, ignoring JSON-format judgment, yields 105/325 passes (32.31\%) versus 39/325 (12.00\%): 66 additional passes. We report this scoring-protocol comparison separately from the strict training comparison, with protocol details in Appendix~\ref{sec:transfer-protocol}.

Gains on REAL, Odysseys-Easy, and WebArena-Verified Overall support the learning utility of environment-interaction data. The before/after comparison measures SFT transfer, rather than isolating individual reconstruction mechanisms.

\paragraph{D. Comparing environment data sources.}
Using the same backbone, training configuration, and evaluation harness, we compare \system{}, WebHarbor, and VeriEnv data at similar trajectory counts (661, 617, and 663). \system{} leads both Rubric Avg and Task Avg on 189 jointly valid Odysseys-All tasks (Table~\ref{tab:data-source-comparison}). Data construction, unequal training-row counts, and validity filtering are detailed in Appendix~\ref{sec:data-source-comparison-protocol}.

\begin{table}[!htbp]
\centering
\begingroup
\small
\setlength{\tabcolsep}{3pt}
\renewcommand{\arraystretch}{1.12}
\begin{tabularx}{\linewidth}{Xrrr}
\toprule
Training data & \shortstack{Rubric\\Avg} & \shortstack{Task\\Avg} & Perfect \\
\midrule
WebHarbor & 6.93 & 7.79 & 0.00 \\
VeriEnv & 6.42 & 7.53 & 0.00 \\
\system{} & \textbf{10.44} & \textbf{11.70} & \textbf{0.53} \\
\bottomrule
\end{tabularx}
\caption{Odysseys-All on 189 jointly valid tasks (\%). Perfect counts: \system{} 1/189; both baselines 0/189. Data and validity definitions: Appendix~\ref{sec:data-source-comparison-protocol}.}
\label{tab:data-source-comparison}
\endgroup
\end{table}

\section{Conclusion}
\label{sec:conclusion}

We present \system{}, an end-to-end automated framework
for transforming real websites into source-grounded,
executable, and reusable Web agent environments.
Unlike human-assisted mirroring approaches, \system{}
automates source exploration, evidence collection,
website reconstruction, independent verification,
and iterative refinement, eliminating the need for
per-site manual implementation and debugging in
our audited construction runs.
By grounding reconstruction in browser-observable
structures, resources, and interaction behaviors,
the framework reduces reliance on incomplete visual
references or manually specified workflows.

Structural, resource, and behavioral evidence guides
reconstruction, while independent verification
directs discrepancies to local repair or targeted
recapture.
This separation enables the system to
distinguish implementation defects from insufficient
source evidence and refine reconstructed environments
accordingly.
The framework constructs 244 website artifacts
across 14 application domains and 53 categories,
supporting 12,984 executed and accepted tasks.
Experiments demonstrate source fidelity,
operational validity, and verification-driven
refinement, while also highlighting the importance
of evaluating functional coverage independently
from visual similarity.
Furthermore, supervised fine-tuning on 5,856
interaction trajectories improves Web agent
performance on REAL, Odysseys-Easy, and
WebArena-Verified, demonstrating the downstream
utility of the constructed environments.

Together, these results establish a practical path
toward scaling real-Web agent environments with
minimal site-specific human engineering.
More broadly, \system{} connects automated website
reconstruction with reproducible interaction and
training-data collection, providing a foundation
for scalable Web agent development and post-training.

\clearpage
\section*{Limitations}
\label{sec:limitations}

\paragraph{Training validation with large-scale models.}
Our supervised fine-tuning experiments with Qwen3.5-9B demonstrate the learning value of interaction data from reconstructed environments for downstream Web agent tasks. However, current training experiments primarily concern the 9B parameter scale. We have not systematically evaluated these data with larger foundation models, different post-training methods, or larger training budgets. Future work can examine relationships among environment scale, data diversity, and model capabilities to assess this approach in larger-scale agent training.

\paragraph{Continuous real-Web updates and environment maintenance.}
Real websites continually change their page structures, resources, and interaction behaviors. \system{} currently reconstructs and verifies source states at particular times, without systematically evaluating the timeliness of local mirrors or long-term maintenance efficiency under continuing source changes. Future research can investigate source-change detection, incremental evidence capture, and local environment updates to reduce maintenance costs while preserving source correspondence.

\paragraph{Copyright, content authorization, and environment distribution.}
High-fidelity reconstruction of real websites involves page content, images, fonts, and branding that may be protected by copyright, trademarks, and terms of service. Public accessibility does not necessarily grant permission to reproduce, modify, or redistribute content. Environment construction, research use, and public release must therefore comply with applicable content authorization and compliance requirements. Future releases can prioritize the general construction framework, evaluation protocols, and de-identified research data, while determining the distribution scope of website mirrors and original resources according to their licenses.

\section*{Ethical Considerations}
\label{sec:ethics}

Exploring, capturing, and reconstructing real websites raises issues concerning terms of service, intellectual property, privacy, and security. The system should process only websites covered by authorization, public licenses, or research agreements, avoid collecting account credentials, cookies, personal information, and payment content, and run released environments in isolated networks and sandboxes. For websites or tasks that could facilitate abusive automation, reconstruction scope, public content, and model capabilities should be restricted.

\bibliography{references}
\appendix
\section{Environment Construction and Quality Evaluation}

The appendices provide protocols, implementation details, per-site results, and source inventories for the completed experiments.

\subsection{Historical Construction and Cost Records}
\label{sec:historical-construction-records}

Construction reports for the historical 85-site subset record 59 passed sites, 15 source-blocked sites, 6 failed sites, and 5 sites with missing reports. These statuses reflect different versions of acceptance criteria, visual algorithms, and thresholds and serve to trace historical construction processes. After initial setup and authorization, this subset contains no recorded per-site manual implementation, debugging, or repair. Conclusions about human intervention and pricing are restricted to this historical subset and the recovered calls.

This construction experiment used the Full Coverage policy. In the recovered SDK records, the Builder and Verifier primarily used \texttt{gpt-5.6-sol} with \texttt{xhigh} reasoning; some Builder calls used \texttt{gpt-6-astra}, so the entire batch did not use a single model.

Recovered model-call and partial cost records cover 84/85 sites. Input tokens, including cached tokens, total 11.207 billion, of which 10.672 billion are cached input tokens; output tokens total 38.4318 million. Recorded, quantifiable model costs total \$220.62, with recorded Builder and Independent Verifier costs of approximately \$142.52 and \$78.11, respectively. Across the 84 sites with records, the mean recorded per-site cost is approximately \$2.63 and the median is approximately \$1.60. Independently rounded components sum to \$220.63, differing from the total by \$0.01 due to rounding. Seven sites additionally contain unpriced Astra calls. These costs therefore provide only a lower bound for recovered, priceable model calls. They exclude unpriced calls, browser execution, and other system overhead, are not extrapolated to the total cost of 244 sites, and are not interpreted as the marginal cost of the Verifier under a controlled ablation.

Costs distinguish system development, initial authorization/setup, per-site implementation/debugging/repair, and research auditing. Neither the 244-site artifact pool nor the historical 85-site subset has a complete attempt cohort under uniform acceptance criteria, so we do not compute a unified construction success rate.

\subsection{Evidence Scope of Independent Evaluation}
\label{sec:result-provenance}
\label{sec:artifact-eligibility}
The categories, screenshot hashes, and audit statuses of the 244-site inventory are archived with the supplementary materials. Automatic metrics for the five sites and 14 states in RQ2-A derive from per-page scoring files and equally weighted per-site aggregation. The six screenshots displayed in Figure~\ref{fig:historical-qualitative} are linked through scoring records and capture hashes; all nine original illustrative screenshots remain archived with the figure's provenance manifest. Historical-reference and mirror screenshots are traced separately from the text extraction used for Text F1; construction-stage checks and the paper's independent measurements are recorded separately.

RQ3 reports paired comparisons on the tasks for which both models obtained valid results and separately reports complete task denominators and infrastructure failures. Evaluation settings, exact counts, and rounding rules appear in Appendix~\ref{sec:transfer-protocol}. Results use descriptive statistics.

Table~\ref{tab:artifact-availability} lists the external artifacts actually included. The four additional sites matched between our artifacts and VeriEnv include artifacts imported from existing directories; their execution results do not retroactively count toward independent SDK construction success. Test adapters connect only execution, measurement, and initial-state recovery interfaces; they neither repair nor simplify the implementations under test. Different subsets are reported separately rather than merged into a ranking.

\begin{table}[!htbp]
\centering

\footnotesize
\setlength{\tabcolsep}{3pt}
\begin{tabularx}{\linewidth}{p{0.22\linewidth}p{0.13\linewidth}X}
\toprule
System & B sites & A states and reference limitations \\
\midrule
WebHarbor & 5/5 & Five sites/14 states; historical Apple and ESPN references have rendering defects \\
VeriEnv & 5/5 & LinkedIn/four states; Coursera is excluded from historical paired scoring; the other three sites lack historical references \\
\bottomrule
\end{tabularx}
\caption{External artifact eligibility for RQ2. A denotes historical paired fidelity; B denotes offline interaction validity.}
\label{tab:artifact-availability}
\end{table}

\subsection{Historical Source-State Fidelity: References and Sensitivity Analysis}
\label{sec:historical-fidelity-details}

\begin{figure*}[!t]
\centering
\includegraphics[width=0.94\textwidth]{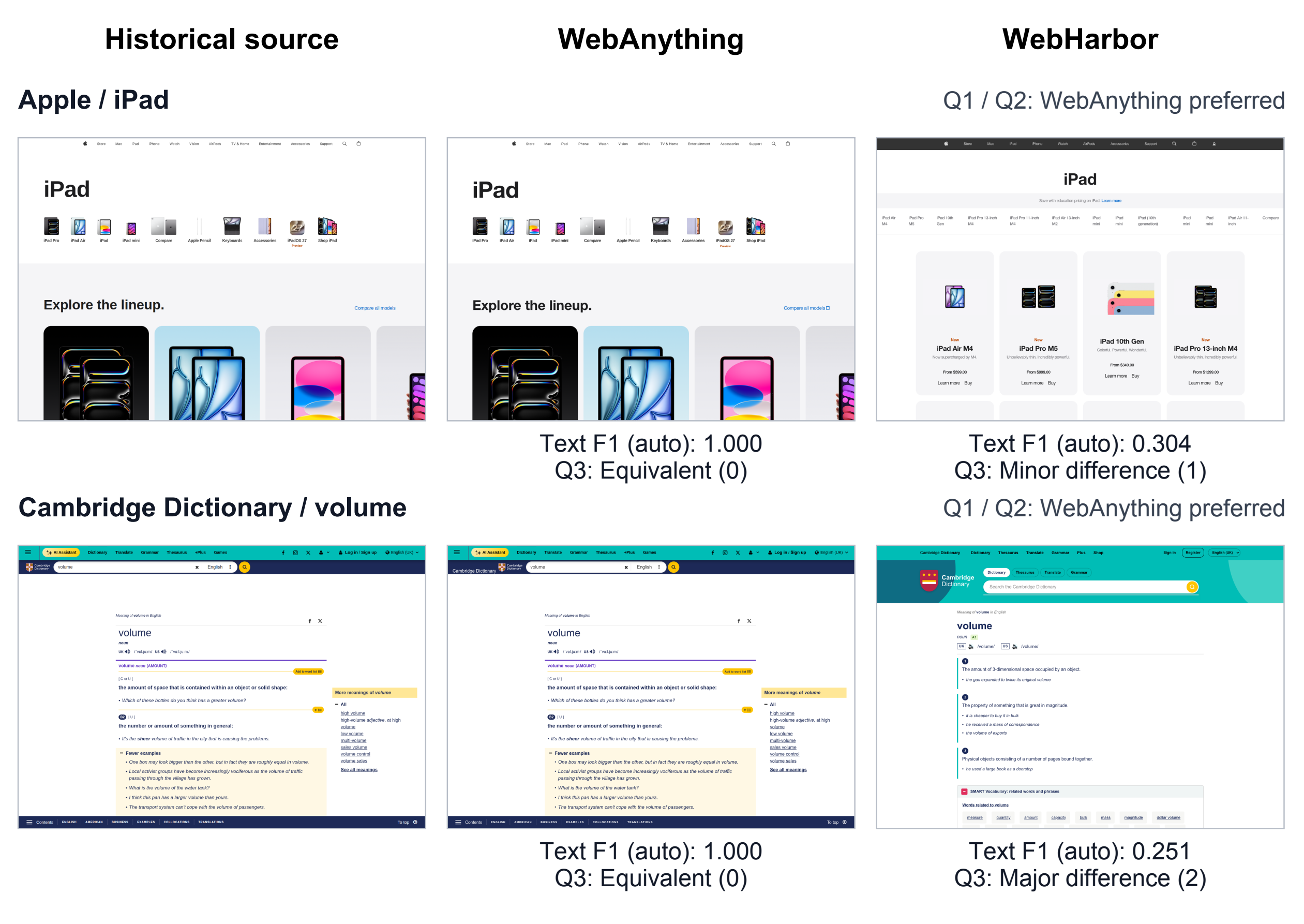}
\caption{Blind-audit page comparisons between historical source references and reconstructed mirrors. Rows show Apple iPad and a Cambridge Dictionary entry. Q1/Q2 indicate overall and information-fidelity preferences; Q3 indicates discrepancy severity: 0 equivalent, 1 minor, and 2 severe. Automated Text F1 is annotated separately from human ratings. The historical Apple reference records a failed font-stylesheet request. Aggregate blind-audit results across all 14 states are reported in Table~\ref{tab:source-fidelity}.}
\label{fig:historical-qualitative}
\end{figure*}

Historical references for Apple and ESPN have rendering defects, and JavaScript rendering conditions differ between source and mirror pages. We retain them in the main results and conduct a sensitivity analysis excluding these two sites: SSIM is 0.8051 for \system{} and 0.6885 for WebHarbor. This analysis does not replace the five-site main results. Historical paired evaluation of VeriEnv covers four LinkedIn states, with Pixel/SSIM of 0.9574/0.8190 and 0.9436/0.7637, respectively; Text F1 is N/A. Its historical references contain personalized Chinese content, whereas some mirrors contain synthetic English data, limiting case interpretation. Coursera is excluded from historical paired scoring, and the other three sites lack historical references. Current-source spot checks are reported separately in Appendix~\ref{sec:current-source-spotcheck}.

Text F1 in Figure~\ref{fig:historical-qualitative} corresponds to the displayed page state rather than the site mean; human labels come from the corresponding frozen blind-audit samples. Different metrics need not agree: for the arXiv abstract page not displayed in the figure, SSIM is 0.631 for WebAnything and 0.655 for WebHarbor. Its WebHarbor 404 state remains in the original screenshots and complete evaluation statistics.

Historical Text F1 uses full-page \texttt{body.innerText}, NFKC and whitespace normalization, and case-sensitive word/punctuation token multisets. Precision and recall use multiset overlap relative to mirror and source token counts. The archived protocol and extraction scripts replay frozen source HTML and resources with JavaScript disabled, retaining historical static normalization; mirror text comes from the earlier JavaScript-enabled captures. Source text is therefore a replay-derived reference, not a contemporaneous live-source text snapshot. This archival rendering asymmetry can affect Text F1 and distinguishes it from matched live rendering; the recorded scores are retained.

\paragraph{Freezing, execution, and denominator rules.}
The resource inventory and environment-quality evaluation materials in this section use records frozen on October 9, 2026: resource scale comes from the frozen 244-site inventory, historical construction/cost audits from the 85-site subset, and quality results from two independent artifact comparisons. The construction-stage Verifier drives repair and recapture according to site contracts. The paper's independent evaluation harness is separate: it neither reads historical verification conclusions nor modifies the artifacts under test, and instead retests each artifact using common metrics. Artifacts, historical references, and runtime configurations are frozen separately. All failures, 404 responses, source-reference defects, blocked tests, and missing references are retained; no composite environment-quality score is constructed.

Historical-state comparisons use the same designated reference, browser, and viewport, recording limitations in the comparability of page states, languages, and JavaScript rendering conditions. The evaluation harness is independent of construction methods and implementation stacks. State recovery uses explicit test adapters rather than requiring external artifacts to adopt \system{}'s Flask/SQLite architecture. Interaction tests disable external network access, use a fixed planned set, and record reasons for unexecuted and failed tests. A historical \texttt{passed} status cannot substitute for retesting.

\subsection{Current-Source Spot Checks}
\label{sec:current-source-spotcheck}

We collected current-source references for five sites and 12 page states frozen in advance. Source access used fresh anonymous sessions with a common $1440\times1000$ viewport, DPR=1, en-US, and UTC. We waited 15 seconds after DOM loading and disabled screenshot animations at the top of the page. Only read-only requests were permitted on source sites; mirrors used isolated copies with external network access disabled. Pixel is defined as $1-\mathrm{MAE}/255$, SSIM uses a uniform RGB $7\times7$ window, and Text F1 measures token overlap in full-page \texttt{body.innerText}. The original protocol and per-site results are preserved in the supplementary materials.

Valid source references cover four Coursera and Weather pages. Three render normally on both sides and enter the paired similarity statistics (Table~\ref{tab:current-source-spotcheck}). The current Coursera source is the learner homepage; the selected WebAnything-side artifact is an older learner homepage, whereas VeriEnv's root page is Coursera for Campus. Catalog data for Python search also differ. Weather's source homepage is localized to Hong Kong, whereas the selected artifact shows New York weather; weather, news, and advertising all vary over time. These spot checks therefore measure descriptive similarity to the pages at capture time and are reported separately from historical-reference evaluation. The selected Coursera/Weather artifacts come from existing site directories, and conclusions concern these artifacts.

\begin{table*}[t]
\centering

\small
\setlength{\tabcolsep}{4pt}
\begin{tabularx}{\linewidth}{p{0.12\linewidth}X>{\centering\arraybackslash}p{0.13\linewidth}ccc}
\toprule
Site & Artifact & Paired/preset states & Pixel & SSIM & Text F1 \\
\midrule
Coursera & Selected pool artifact & 2/2 & 0.7017 & 0.4742 & 0.4236 \\
 & VeriEnv & 2/2 & 0.7466 & 0.5338 & 0.1928 \\
Weather & Selected pool artifact & 1/2 & 0.6373 & 0.4386 & 0.1955 \\
 & VeriEnv & 1/2 & 0.5188 & 0.4018 & 0.1396 \\
\midrule
Apartments & Both; source 403 & 0/2 & N/A & N/A & N/A \\
Discogs & Both; source 403/access challenge & 0/2 & N/A & N/A & N/A \\
LinkedIn & Both; pre-login states & 0/4 & N/A & N/A & N/A \\
\bottomrule
\end{tabularx}
\caption{Current-source spot checks. Paired/preset states retain all planned states; means use the same pages that render normally on both sides. Source unavailability and mirror runtime failures are reported separately. Product scope, location, and live-content conditions follow the protocol in this section.}
\label{tab:current-source-spotcheck}
\end{table*}

Two pages each on Apartments and Discogs returned 403 responses or access challenges. Four LinkedIn pages returned login or public landing pages that differed from the mirrors' post-login states. These eight states remain classified as source-unavailable. The Weather New York page rendered normally in the selected WebAnything-side artifact, whereas VeriEnv returned 500 offline because of an external API dependency. This runtime failure remains in the planned-state records, and paired similarity means use only the homepage. This check evaluates page presentation and runtime availability; interaction-test results remain those of the independent RQ2-B records.

\subsection{Website Stratification and Coverage Statistics}
\label{sec:coverage-protocol}

The resource inventory is deduplicated by unique site identifiers and contains 244 artifacts. The first- and second-level usage-category fields correspond to 14 application domains and 53 fine-grained categories, respectively; source hostnames and application domains are recorded separately. Categories retain the initial resource-table annotations, and the pool is not a preregistered random sample with fixed quotas. Site aliases, multiple snapshots, or candidate versions do not count as additional sites. Usage categories in RQ1, tested capabilities in RQ2-B, and actual trajectory visitation scopes are counted separately and are not interchangeable.

\subsection{Presentation Diagnostics and Quality Depth of the Resource Pool}
\label{sec:pool-quality-depth}

All 244 startup records in the frozen inventory contain HTTP 200 responses and homepage screenshots, but these do not represent complete functional acceptance. The 113 additionally reviewed sites comprise 70 without confirmed anomalies, 9 with flagged presentation issues, 26 with signals awaiting diagnosis, and 8 other special records. The remaining 131 sites received no additional review.

SHA-256 and 256-bit difference-hash diagnostics across all frozen screenshots found byte-identical blank homepage screenshots for Cohere and United Airlines. Their historical records correspond to script execution failure and a JavaScript chunk 404, respectively. Fifteen screenshot pairs with difference-hash distance at most 12 are used to locate low-content pages, not to identify duplicate implementations.

The historical Best Buy artifact contains 184 programmatically generated SVGs, and its homepage explicitly uses locally synthesized product and store data. This confirms that some colored blocks are synthetic resources rather than loading failures. Fictional business data do not themselves imply functional invalidity, nor do they establish the pool-wide prevalence of templated implementations. Presentation diagnostics are reported separately from source fidelity and interaction tests; test results for other candidate versions are not reassigned to the current resource pool.

\subsection{Interaction Tests and State Recovery}
\label{sec:validity-probe-details}

\begin{table*}[t]
\centering
\begingroup
\renewcommand{\theHtable}{full-artifact-validity}

\footnotesize
\setlength{\tabcolsep}{3pt}
\begin{tabularx}{\linewidth}{p{0.20\linewidth}*{3}{>{\centering\arraybackslash}X}ccc}
\toprule
System & \shortstack{Capability\\coverage} & \shortstack{Executed-test\\success rate} & \shortstack{All-test\\pass rate} & Startup & Persistence & \shortstack{Initial-state\\restoration} \\
\midrule
\multicolumn{7}{l}{\textit{Five-site WebHarbor comparison}} \\
\system{} & 15/20 (75\%) & 19/20 (95\%) & 19/30 (63.3\%) & 5/5 & 1/1 & 1/1 \\
WebHarbor & 16/20 (80\%) & 23/23 (100\%) & 23/30 (76.7\%) & 5/5 & 2/3 & 3/3 \\
\midrule
\multicolumn{7}{l}{\textit{Five-site VeriEnv comparison}} \\
Selected pool artifacts$^{a}$ & 22/22 (100\%) & 29/31 (93.5\%) & 29/33 (87.9\%) & 5/5 & 6/6 & 5/5 \\
VeriEnv & 20/22 (90.9\%) & 21/28 (75.0\%) & 21/33 (63.6\%) & 5/5 & 4/4$^{b}$ & 5/5 \\
\bottomrule
\end{tabularx}
\par\vspace{3pt}
\begin{minipage}{\linewidth}\footnotesize
$^{a}$ The four added sites include artifacts imported from existing directories; comparisons use the frozen environments.
$^{b}$ Persistence 4/4 refers to executed checks; two additional planned checks were not executed.
\end{minipage}
\caption{Interaction validity of selected frozen environments. WebHarbor and VeriEnv are tested on their respective five-site subsets. We report capability coverage, success rate over executed tests, and pass rate over all predefined tests. External network access is disabled; explicit test adapters restore state.}
\label{tab:full-environment-validity}
\endgroup
\end{table*}

Artifact-level tests use fixed site-capability sets covering clicks, text entry, retrieval, navigation, form feedback, and state modification. Capability coverage uses the number of planned capabilities as its denominator; supported and actually tested capabilities count in its numerator even when execution fails. Executed-test success rate uses only executed tests as its denominator, whereas all-planned-test pass rate retains unsupported and blocked items. Image-resource errors are reported separately rather than mixed into interaction-test success rates.

All-planned-test results for the five VeriEnv sites are LinkedIn 6/9 vs.\ 7/9, Apartments 6/6 vs.\ 4/6, Coursera 6/6 vs.\ 4/6, Discogs 6/6 vs.\ 3/6, and Weather 5/6 vs.\ 3/6 (selected pool artifacts vs.\ VeriEnv), totaling 29/33 vs.\ 21/33.

In the five-site WebHarbor subset, WebHarbor achieves higher coverage and interaction-test completion rates. \system{}'s Cambridge lacks explicit no-results feedback for nonexistent words. WebHarbor's Apple login session becomes invalid after restarting, but shopping-bag state persists. Ten distinct image-resource errors on ESPN are recorded separately.

Initial states are restored through explicit test adapters, and native reset-API support is recorded separately. Persistence/recovery checks are performed only after the corresponding state modification succeeds. VeriEnv's executed persistence checks are 4/4, with two additional planned checks unexecuted; these remain in the planned set. This evaluation reports adapter-level recovery results and does not imply that all artifacts have native reset interfaces.

\section{System Implementation and Verification Feedback}

\subsection{Propagation of Evidence, Candidates, and Feedback}

The Builder reads existing source materials, candidates, and the latest repair package in the workspace, preserving valid captures and implementations when resuming execution. The independent verification runner reads construction reports and evidence directories and returns verification reports and repair requests. The outer controller records execution statuses and sends failed reports back to the Builder; source blocking and authorization requirements are handled separately. Task generation, interaction collection, and evaluation use frozen environments.

\subsection{Version Association}

Sites, source snapshots, candidates, and release packages have separate stable identifiers. Version associations for tasks, trajectories, and evaluation results should identify specific implementations, resources, initial data, and runtime configurations rather than relying solely on site names. Training and evaluation protocols and data-overlap checks appear in Appendix~\ref{sec:transfer-protocol}.

\subsection{Site Contracts and Verification-Report Fields}

The inspected contracts associate sites, coverage configurations, viewports, page paths, workflows, and acceptance requirements, with optional interaction contracts and state assertions. Specific checks are determined by the corresponding execution reports.

Independent verification reports record conclusions, candidate digests, page and workflow checks, source limitations, and evidence. Repair packages associate issue locations, expected and observed outcomes, corresponding evidence, and repair suggestions. The model determines diagnostic actions using the reports and existing source evidence.

\subsection{Construction-Stage Verifier Implementation and Versions}
\label{sec:verifier-implementation-audit}

The construction-stage Verifier uses site contracts, model-executed browser tests, and site-specific assertions. Visual comparison may combine screenshots, DOM, element geometry, and auxiliary scripts. Historical versions did not use a unified SSIM, LPIPS, or L1--L4 composite score. Fidelity v19 introduced normalized RGB pixel similarity, but algorithms and thresholds still differ across sites.

Reset prechecks include programmatic hard checks dependent on \system{}'s Flask/SQLite architecture; they are not stack-independent external-baseline scorers. Earlier runners did not enforce positive- and negative-case calibration acceptance criteria for evaluators. Task-evaluation calibration therefore requires independent evidence.

Reports preserve concrete check evidence, with top-level conclusions distinguishing passed, failed, and source-blocked outcomes. The outer runner consumes reports rather than recomputing a unified four-layer score. The historical artifact pool contains candidates from different versions with passed, failed, source-blocked, and missing-report statuses. The independent evaluation harness retests frozen artifacts without reading these historical verification conclusions.

\subsection{Resumption Prompt for the Explorer and Builder}
\label{sec:coupled-builder-prompt}

The Builder Agent directly receives a real website URL and manages exploration, evidence collection, and local reconstruction within the same context. The historical resumption template requires preserving existing source evidence, candidate implementations, and repair records, reading the latest check feedback, and continuing repairs from failed or unfinished items. It returns to the source for targeted recapture only when existing evidence cannot answer a concrete question. Its final submission comprises the candidate implementation, local resources, and a self-check report containing actual check results and limitations. This template resumes checkpoints after timeouts; it is not the sole initial prompt shared by all construction batches.

The complete historical English template appears in the separate prompt supplement; this description does not replace the original instructions.

\subsection{Independent Final Verifier Prompt}
\label{sec:final-verifier-prompt}

The Independent Verifier inspects frozen candidates in a separate context without modifying the Builder's candidate code. It performs actual browser checks based on source evidence and site contracts, covering applicable pages, controls, resources, state persistence, and recovery, and produces feedback containing concrete failure locations, expected outcomes, observed outcomes, and evidence. Stateful checks modify only local runtime states and restore initial states before finishing. Candidates modified by the Builder must be frozen again and independently reverified.

The English template and unexpanded input/output variables are preserved in \nolinkurl{supplementary-prompts.tex}. Internal historical names are anonymized as construction-system identifiers, retaining the original instructions and system attribution. The template's construction-acceptance viewport is $1920\times992$, distinct from the $1440\times1000$ viewport used for independent quality evaluation. Executed historical checks are documented in the corresponding records.

\subsection{Evidence Definitions for Verification Decoupling and Feedback Analysis}
\label{sec:verification-analysis-details}

\begin{table}[!t]
\centering

\begingroup
\small
\setlength{\tabcolsep}{4pt}
\renewcommand{\arraystretch}{1.12}
\begin{tabularx}{\columnwidth}{>{\raggedright\arraybackslash}X>{\centering\arraybackslash}p{0.17\columnwidth}>{\centering\arraybackslash}p{0.19\columnwidth}}
\toprule
Independent outcome & Pairs & Share \\
\midrule
Passed & 56 & 25.57\% \\
\textbf{Failed} & \textbf{139} & \textbf{63.47\%} \\
Source blocked & 24 & 10.96\% \\
\midrule
\textbf{Of these: evidence-supported discrepancies} & \textbf{39} & \textbf{17.81\%} \\
\bottomrule
\end{tabularx}
\par\vspace{3pt}
\begin{minipage}{\columnwidth}\footnotesize
All shares use 219 builder-pass pairs as the denominator; the 39 pairs span 22 sites. Candidate identity, check scope, or runtime conditions are not strictly matched in every pair. Only one check pair strictly matches both candidate and semantic target.
\end{minipage}
\caption{Independent verification of submitted candidates with passing builder self-checks. The first three outcomes are mutually exclusive; the last row is a subset of failures supported by concrete defect evidence.}
\label{tab:independent-verification}
\endgroup
\end{table}

The analysis uses archived construction, self-check, independent-verification, and repair records without rerunning construction or checks. The 229 pairs retain actual associations between construction and verification reports. Builder-pass/Verifier-fail status discrepancies occur in 139 pairs, of which 39 have supporting evidence of concrete check targets and defects. Supporting evidence includes one strictly candidate-matched subcheck and 38 non-strict pairs; sites can overlap across evidence levels.

Among all pairs, three pairs across two sites have identical observed hashes and explicitly identical algorithms, but only one resource subcheck for Stashmedia also satisfies the same semantic target and actual-defect criteria. Other pairs may differ in candidate identity, coverage, protocol, or access conditions. The absence of a target status discrepancy does not prove self-check completeness. Report-level repair chains across 46 sites follow existing historical records and are defined by traceable feedback reading, modification, new digests, and subsequent reverification; they do not count resolution of all initial issues.

The issue table contains 1,007 records, of which 49 received item-level semantic adjudication. The 18 confirmed records jointly demonstrate feedback execution, candidate modification, and a corresponding semantic-reverification endpoint. A later overall pass alone does not establish issue closure. Remaining unadjudicated or incomparable records cannot be treated as repair failures. The 18 records include shared mechanisms, repeated manifestations, and recurrences and are not treated as independent bugs or independent statistical samples. Existing records support actual recapture on two sites, but do not establish closure of the original evidence gaps targeted by recapture under comparable checks; we therefore do not report causal effects of recapture.

\subsection{Evidence for Verification and Feedback Cases}
\label{sec:verification-case-evidence}

\begin{figure*}[tbp]
\centering
\includegraphics[width=\textwidth]{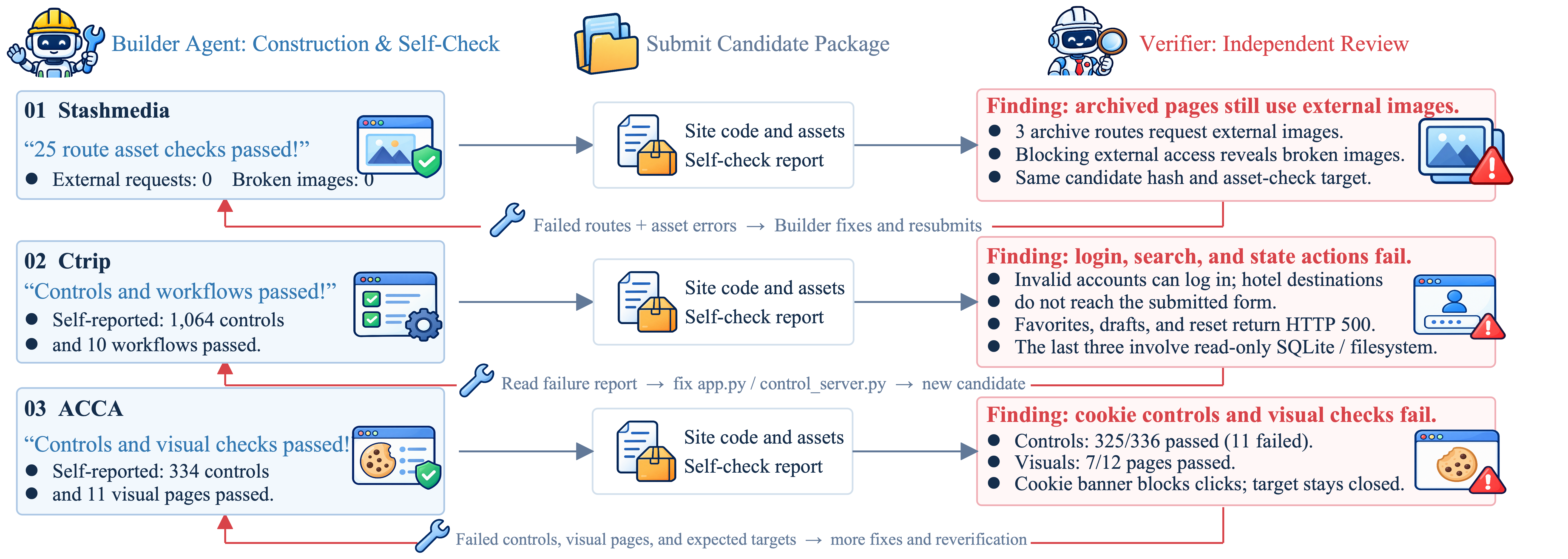}
\caption{Historical verification cases. Stashmedia reveals a same-candidate resource-check omission; Ctrip traces feedback-driven revision and issue-level independent reverification; ACCA illustrates additional control and visual defects. Evidence scopes are specified in Appendix~\ref{sec:verification-case-evidence}.}
\label{fig:verification-feedback-cases}
\end{figure*}

\paragraph{Stashmedia.}
The Builder reports no external requests or broken images in resource checks across 25 routes. The Independent Verifier identifies external image requests, broken images, and console errors on three archive routes. Both sides use the same candidate-digest algorithm, with matching digests and file counts (626), supporting a same-candidate resource-check comparison. The conclusion is restricted to this subcheck; identical waiting and browser-execution conditions have not been fully established.

\paragraph{Ctrip.}
Repair requests record five issues: invalid login, destination queries, favorites, booking drafts, and reset. Construction event streams confirm that feedback was read and that application and runtime-control implementations were subsequently modified. The updated candidate has corresponding semantic-reverification pass records in independent workflow and initial-state recovery checks. The initial failures of favorites, drafts, and reset involve a read-only filesystem/SQLite, so these three records are classified as runtime-environment repairs, and the first two as local-implementation repairs. The initial failing round is outside the strictly associated 229-pair cohort and is reported separately as a repair case, not counted in its 22-site evidence-supported discrepancy cohort.

\paragraph{Figma.}
An authorized source capture established a canvas-anchored comment composer. Verification feedback identified mismatched placement and appearance. The event stream records feedback consumption and changes to the editor markup, styles, and interaction code. Independent targeted checks on the updated candidate confirmed the composer's geometry and visibility. This example illustrates source-supported Local Repair, without attributing resolution to Targeted Recapture.

\paragraph{ACCA.}
The Builder reports passing checks for 334 controls and 11 visual pages. Independent checks record 325 of 336 controls passing and seven of 12 visual pages passing; a cookie banner intercepts 11 cookie-policy controls, preventing their intended destinations from opening. This is a supported historical report discrepancy, not a strictly candidate-matched comparison: the candidate-digest algorithms and check scopes differ. Later popup checks do not establish correct navigation to the expected destination, so these records are not counted among the confirmed issue resolutions.

\section{Task Supply and Data Statistics}

\subsection{Interfaces for Downstream Task Generation, Evaluation, and Collection}
\label{sec:environment-task-usage}

Task and training-data construction use released environments. This section describes the implementation interfaces for task specifications, runtime recovery, evaluation, and data binding.

\paragraph{From environment capabilities to tasks.}
The task pipeline reads the pages, visible objects, native operations, and seed data of a specific mirror to formulate goals such as retrieval, filtering, detail verification, comparison and aggregation, and state modification. Existing planning materials record whether the required facts reside in lists, detail pages, or other pages, and organize applicable operations into task specifications. Natural-language instructions express user goals, while answers, evidence supporting the operations, and evaluation materials remain hidden. The executable scope of a task is bounded by the capabilities and initial state actually supported by the environment.

\paragraph{Initial states and success conditions.}
Public task records contain task identifiers, instructions, sites, starting URLs, and login and reset requirements. Task specifications record capability tags, action and step budgets, and successful and unsuccessful terminal states. Hidden specifications associate required objects, query parameters, source evidence, expected results, and output contracts. Success conditions for information tasks may specify fields, sets, orderings, or comparison results in normalized answers. State-changing tasks check application states after saving, creating, modifying, or restoring objects; tasks admitting multiple solutions use sets of state predicates. When necessary, fictional test records are prepared in private copies of native data structures, while original environment assets remain read-only. Task answers are grounded in the task's initial data. Task text or a nonempty answer file alone does not establish evaluability: corresponding executable checks must exist.

\paragraph{Evaluator construction and execution-based verification.}
During generation, site-specific scripts first check native routes, evidence visibility, and applicable answer, state, and reset conditions. A visual agent then receives instructions and observations in the initial state of the corresponding mirror through the fixed harness and performs browser operations. An independent task evaluator checks the final answer against hidden success conditions and determines goal completion using the terminal DOM, screenshots, and applicable state evidence. Model-based evaluation is used for applicable semantic conditions and is separated from agent execution. Tasks with evaluation materials become end-to-end accepted tasks only after actual interaction and outcome checks, rather than receiving a pass solely because specifications exist.

\paragraph{Recovery, binding, and training entry points.}
Task execution packages locate native implementations, immutable seeds, and starting routes through runtime descriptions. When login is required, they restore fictional seed identities while isolating real user sessions. Before task execution, mutable databases are restored from seeds, and site-specific reset and startup checks verify applicable states. Tasks whose runtime prerequisites are not met are recorded as blocked. Task identifiers, source-data versions, and specification digests connect public tasks to hidden evaluation materials. Collection packages record implementations, resources, initial data, runtime configurations, and archive digests to associate the complete environment identity.

The pipeline separately records task generation, evaluator construction, execution-based acceptance, trajectory collection, and training-sample export. Saved successful trajectories are used for training after quality filtering, format conversion, and step-level sample export. The 6,172 accepted tasks in the earlier source collection, the 5,856 trajectories used for SFT, the 5,167 unique training task IDs obtained by deduplicating task IDs in the training manifest, and the 60,517 step-level samples are counted in their respective units. The relationship between the separate batch of 6,812 new tasks across 229 sites and the 74 source sites follows the set definitions in the experimental setup.

\subsection{Diversity and Interaction Complexity of Accepted Tasks}
\label{sec:verified-task-diversity}

This section reports statistics for the 6,172 accepted tasks in the post-training source collection, covering 74 site labels. Agents executed these tasks in the environments, and task evaluators confirmed success. This collection and the separate collection of 6,812 newly generated tasks across 229 sites have no task-level containment relationship; the 74 source sites are a subset of the 229 sites. The 6,172 source tasks, 5,856 training trajectories, 5,167 unique training task IDs obtained by deduplicating task IDs in the training manifest, and 60,517 step-level samples are distinct counting units. Overlapping site names do not establish task-level or environment-version correspondence.

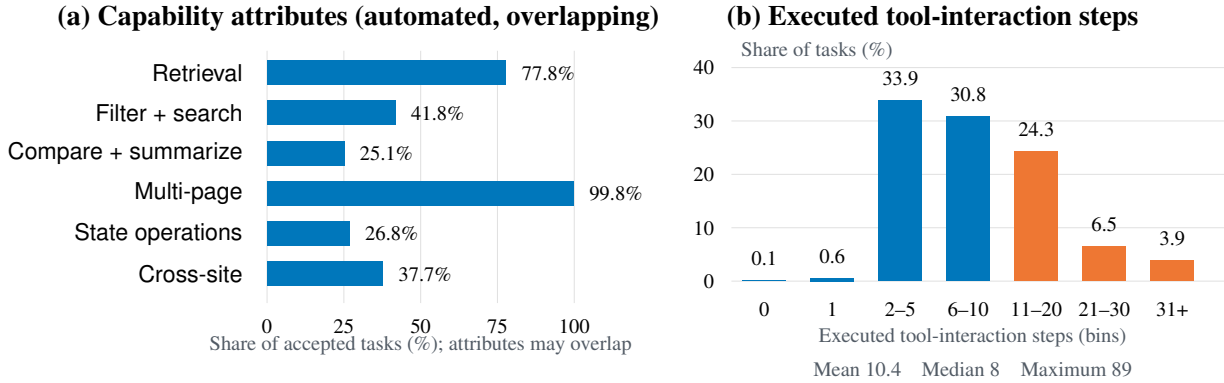
\begin{figure*}[tbp]
\centering
\resizebox{\textwidth}{!}{
\begin{tikzpicture}[x=1cm,y=1cm,font=\sffamily\fontsize{9}{11}\selectfont]
\definecolor{taskblue}{HTML}{0077BB}
\definecolor{taskorange}{HTML}{EE7733}
\definecolor{taskgrey}{HTML}{525B65}
\path[use as bounding box] (-.05,-.75) rectangle (16.4,5.15);
\node[font=\sffamily\bfseries\fontsize{11}{13}\selectfont] at (8.15,4.9)
  {6,172 accepted tasks\quad/\quad74 sites\quad/\quad14 task domains};
\node[anchor=west,font=\bfseries] at (0,4.25) {(a) Capability attributes (automated, overlapping)};
\node[anchor=west,font=\bfseries] at (9.15,4.25) {(b) Executed tool-interaction steps};
\foreach \tick in {0,25,50,75,100}{
  \pgfmathsetmacro{\xt}{3.0+.042*\tick}
  \draw[black!12,line width=.35pt] (\xt,.35)--(\xt,3.8);
  \node[font=\fontsize{8}{10}\selectfont,anchor=north] at (\xt,.27) {\tick};
}
\foreach \lab/\pct/\cnt/\yy in {
  Retrieval/77.8/4802/3.5,
  Filter + search/41.8/2581/2.95,
  Compare + summarize/25.1/1549/2.4,
  Multi-page/99.8/6157/1.85,
  State operations/26.8/1655/1.3,
  Cross-site/37.7/2326/.75}{
  \pgfmathsetmacro{\xe}{3.0+.042*\pct}
  \node[anchor=east] at (2.8,\yy) {\lab};
  \fill[taskblue] (3.0,\yy-.17) rectangle (\xe,\yy+.17);
  \node[anchor=west,font=\fontsize{8}{10}\selectfont] at (\xe+.08,\yy) {\pct\%};
}
\node[text=taskgrey,font=\fontsize{8}{10}\selectfont] at (5.1,-.23) {Share of accepted tasks (\%); attributes may overlap};
\foreach \tick in {0,10,20,30,40}{
  \pgfmathsetmacro{\yt}{.65+.073*\tick}
  \draw[black!12,line width=.35pt] (9.35,\yt)--(16.1,\yt);
  \node[anchor=east,font=\fontsize{8}{10}\selectfont] at (9.25,\yt) {\tick};
}
\node[anchor=west,font=\fontsize{8}{10}\selectfont,text=taskgrey] at (9.35,3.8) {Share of tasks (\%)};
\foreach \lab/\pct/\xx/\col in {
  0/0.1/9.8/taskblue,
  1/0.6/10.73/taskblue,
  2--5/33.9/11.66/taskblue,
  6--10/30.8/12.59/taskblue,
  11--20/24.3/13.52/taskorange,
  21--30/6.5/14.45/taskorange,
  31+/3.9/15.38/taskorange}{
  \pgfmathsetmacro{\yt}{.65+.073*\pct}
  \fill[\col] (\xx-.30,.65) rectangle (\xx+.30,\yt);
  \node[anchor=south,font=\fontsize{8}{10}\selectfont] at (\xx,\yt+.07) {\pct};
  \node[anchor=north,font=\fontsize{8}{10}\selectfont] at (\xx,.50) {\lab};
}
\node[font=\fontsize{8}{10}\selectfont,text=taskgrey] at (12.65,-.12) {Executed tool-interaction steps (bins)};
\node[font=\fontsize{8}{10}\selectfont,text=taskgrey] at (12.65,-.55) {Mean 10.4\quad{}Median 8\quad{}Maximum 89};
\end{tikzpicture}}
\caption{Diversity and interaction complexity of 6,172 accepted source tasks spanning 74 site labels and 14 task application domains. (a) Six automated, overlapping capability attributes derived from instructions, trajectories, and website information, each using 6,172 as the denominator. (b) Grouped distribution of executed tool-interaction steps; approximately 34.7\% of tasks exceed 10 steps.}
\label{fig:task-diversity-complexity}
\end{figure*}

\paragraph{Capability attributes.}
The labels in Table~\ref{tab:verified-task-capabilities} are automatically generated from instructions, execution trajectories, and website information. They may overlap and are not human gold-standard annotations. Task counts are computed per attribute, and site counts reflect the sites associated with that attribute. Neither column can be summed across rows to obtain a total number of unique objects. The multi-page attribute retains its automatic-label definition and is not directly equivalent to the number of unique paths after query parameters are removed.

\begin{table}[!htbp]
\centering

\small
\setlength{\tabcolsep}{3pt}
\begin{tabularx}{\linewidth}{Xrrr}
\toprule
Task attribute & Tasks & Share & Sites \\
\midrule
Information retrieval & 4,802 & 77.8\% & 74 \\
Filtering and search & 2,581 & 41.8\% & 72 \\
Comparison and aggregation & 1,549 & 25.1\% & 68 \\
Multi-page tasks & 6,157 & 99.8\% & 74 \\
State operations & 1,655 & 26.8\% & 68 \\
Cross-site tasks & 2,326 & 37.7\% & 28 \\
\bottomrule
\end{tabularx}
\caption{Automatically assigned multi-label capability statistics for 6,172 accepted tasks.}
\label{tab:verified-task-capabilities}
\end{table}

\paragraph{Observed interaction complexity.}
Table~\ref{tab:verified-task-complexity} uses actual execution trajectories rather than planning estimates. Tool-interaction steps count tool operations during execution; URL page states retain query parameters, whereas page paths discard them. These three statistics are not interchangeable. Other URL-normalization and repeated-state handling rules must accompany the statistical scripts; we do not infer unreported deduplication rules from aggregates. Step-count bins appear in Table~\ref{tab:verified-task-step-bins}: 2,139/6,172 tasks exceed 10 steps (approximately 34.7\%), and 641/6,172 exceed 20 steps (approximately 10.4\%). These values demonstrate variation in interaction depth, not that all tasks are long-horizon. Zero-step tasks remain in the total denominator and are not excluded post hoc.

\begin{table*}[!t]
\centering

\small
\setlength{\tabcolsep}{7pt}
\begin{tabularx}{\linewidth}{Xrrrrrr}
\toprule
Metric & Mean & Median & P75 & P90 & P95 & Maximum \\
\midrule
Observed tool-interaction steps & 10.4 & 8 & 13 & 21 & 28 & 89 \\
Observed URL page states & 5.5 & 5 & 7 & 9 & 12 & 63 \\
Page paths excluding query parameters & 4.7 & 4 & 6 & 8 & 10 & 62 \\
\bottomrule
\end{tabularx}
\caption{Observed interaction complexity of accepted tasks. Tool operations, URL states, and page paths are counted separately from actual trajectories.}
\label{tab:verified-task-complexity}
\end{table*}

\begin{table}[!htbp]
\centering

\small
\setlength{\tabcolsep}{8pt}
\begin{tabular}{lrr}
\toprule
Tool-interaction steps & Tasks & Share \\
\midrule
0 & 4 & 0.1\% \\
1 & 38 & 0.6\% \\
2--5 & 2,092 & 33.9\% \\
6--10 & 1,899 & 30.8\% \\
11--20 & 1,498 & 24.3\% \\
21--30 & 399 & 6.5\% \\
31 or more & 242 & 3.9\% \\
\midrule
Total & 6,172 & --- \\
\bottomrule
\end{tabular}
\caption{Bins of observed tool-interaction steps. Rounded shares may not sum to 100\%.}
\label{tab:verified-task-step-bins}
\end{table}

\subsection{New Task Batch and Acceptance Definitions}
\label{sec:task-supply-snapshot-protocol}

The new batch and the earlier collection of 6,172 post-training source tasks are independent sets. The earlier 74 source sites are a subset of the 229 sites in the new batch. Within the 244-site artifact pool, 229 sites generated 6,812 tasks that completed visual-agent execution and corresponding task-evaluator acceptance. Of these, 217 sites reached 30 tasks each, 12 generated 1--29 tasks, and 15 have no generated tasks. Across all 244 sites, the median is 30 tasks per site, the range is 0--30, and the mean is 27.9. Target-quota completion is 6,812/7,320 (93.1\%). These counts describe accepted outputs, not the success rate of all proposals or execution attempts. The newer collection is intended for subsequent agent training and was not used in the 5,856-trajectory SFT experiment reported here.

The supporting-materials snapshot of October 10, 2026, contains 6,812 public task records and matching metadata with unique task identifiers. It also records 6,782 matched hidden-criterion identifiers and 6,540 runtime-description files. Runtime descriptions support environment setup and execution, whereas task acceptance is established by agent execution and corresponding evaluator checks. These file-level counts summarize archived supporting materials and are not used as a task-to-runtime-file completion ratio or a count of independent environment-validation outcomes.

Task generation, evaluation materials, script checks, end-to-end acceptance, trajectory collection, and training export are counted separately. The SFT collection is specified in Section~\ref{sec:downstream-utility}; trajectory-length statistics use observed actions.

\section{Human Blind Audit: Protocol and Results}
\label{sec:human-state-audit}

The human blind audit evaluates historical source-state fidelity in RQ2-A. All five sites and 14 page states in the WebHarbor-matched historical-state subset are included: Apple, arXiv, Cambridge Dictionary, ESPN, and Hugging Face Papers. There is no sampling or selection based on automatic scores. Each sample simultaneously displays the historical source reference, anonymous mirror A, and anonymous mirror B, with A/B order randomized per sample. Three annotators participate, forming final labels through discussion. Annotators do not know system identities or see Pixel, SSIM, or Text F1.

\paragraph{Rating questions.}
Q1 (overall source-state fidelity) asks which mirror is closer overall to the designated historical source page, with choices A better/tie/B better. Q2 (information fidelity) asks which mirror more completely preserves visible information that may affect agent observation or decisions, using the same choices. Q3 (agent-relevant discrepancy severity) separately rates each mirror against the historical source reference: 0 denotes equivalent visible states, 1 minor differences in visible information or interaction cues, and 2 severe discrepancies in those observations or cues; lower is better. The target is always the designated historical source state. Annotators are not asked to infer functional correctness for operations that were not executed.

\paragraph{Consensus annotations.}
All wins/ties/losses are computed for \system{} relative to WebHarbor. Q1 records 12 wins, 2 ties, and 0 losses, with a win rate of $12/14=85.7\%$. Q2 records 10 wins, 3 ties, and 1 loss, with a win rate of $10/14=71.4\%$. Both rates use all 14 pages as the denominator; ties are not removed. The sole WebHarbor win in Q2 corresponds to ESPN.

For Q3, \system{}'s equivalent/minor/severe distribution is 12/14, 2/14, and 0/14, with mean severity $(0\times12+1\times2+2\times0)/14=0.143$ (rounded to three decimals). WebHarbor's distribution is 0/14, 8/14, and 6/14, with mean severity $(0\times0+1\times8+2\times6)/14=1.429$. Severe-discrepancy shares are 0\% and 42.9\%, respectively. These quantities are reported separately, not combined into a new overall fidelity score.

\paragraph{Labels and interpretation scope.}
We report consensus labels formed through discussion among three annotators and do not use them to compute independent inter-annotator agreement. Human judgments measure visible differences in these 14 designated historical-reference states; limitations arising from source versions, Apple/ESPN reference-rendering defects, and JavaScript rendering conditions remain applicable. Interaction validity is measured in RQ2-B; the reliability of Verifier conclusions is a separate calibration objective.

\section{Training and Evaluation Configuration}

\subsection{RQ3: Protocol for Post-Training Utility}
\label{sec:transfer-protocol}

We compare the Qwen3.5-9B baseline with the model supervised-fine-tuned on environment interaction data, using the step-473 checkpoint at the fixed training endpoint.

\paragraph{Agent harness and execution settings.}
The fixed harness is shared across browser-based data collection and downstream evaluations. Within each benchmark, the baseline and SFT models use the same harness. Its hybrid browser-interaction setting combines visual and DOM-assisted observations with semantic and coordinate-based actions. Observations include current-viewport screenshots, with numbered Set-of-Mark (SoM) annotations when available, and compact text snapshots derived from the DOM and accessibility structure. These snapshots provide element references (\texttt{ref}), roles, labels, states, and the page URL and title, rather than raw HTML. Image inclusion is subject to the image budget.

The action space exposes 25 tools: reference-based clicking, hovering, form filling, selection, checking, and dragging; viewport-coordinate clicking, double-clicking, and dragging; text input, key presses, and scrolling; and navigation, tab management, screenshot capture, and text reading. Runtime coordinates use CSS pixels relative to the current viewport; image coordinates in the training exports are normalized to 1000.

Odysseys-Easy uses this hybrid setting, with \texttt{gpt-5.6-sol} as the judge and \texttt{max\_images=40}. The execution limits are 30 model turns and 10,800 seconds per task for REAL, and 100 model turns and 5,400 seconds per task for Odysseys-Easy. These budgets follow the corresponding runtime configurations; execution retries and concurrency are preserved separately in runtime records.

\paragraph{Training inputs and recipe.}
The task configuration actually reads the non-thinking input file, whose independently counted line total is 60,517. The merged data manifest records 5,856 trajectories: 4,651 from the historical batch and 1,205 from the incremental batch. Deduplication by task ID in the training manifest yields 5,167 unique training task IDs across 74 site labels. Versions and format variants of the same task are retained, so trajectory counts are not counts of semantically unique tasks. Filtering requires task-evaluation success and a mean score of at least 4 across process, completion, and efficiency; it does not require each dimension to be at least 4. Only the final agent turn in each sample is supervised.

\begin{table*}[t]
\centering

\small
\renewcommand{\arraystretch}{1.15}
\setlength{\tabcolsep}{6pt}
\begin{tabularx}{\linewidth}{p{.24\linewidth}X}
\toprule
Item & Recovered configuration \\
\midrule
Training method and parameter scope & Supervised fine-tuning; trainable language model, frozen vision encoder and projection layer; BF16 \\
Training-data counts & 6,172 accepted source tasks; 5,856 training trajectories; 5,167 unique training task IDs; 60,517 step-level samples \\
Samples and checkpoint & 60,517 step-level samples; step-473 checkpoint; approximately two nominal data passes \\
Optimizer & Adam with decoupled weight decay $10^{-4}$; $\beta_1=0.9$, $\beta_2=0.95$, $\epsilon=10^{-8}$ \\
Learning rate and schedule & $5\times10^{-6}$, minimum 0; cosine decay; 24 warmup steps \\
Batch size and randomness & Global batch size 256, microbatch size 1; random and mixing seeds both 42 \\
Sequence handling & Maximum 262,144; no packing, no forced padding to the maximum, truncation for overlength sequences \\
Image configuration & Minimum 65,536 pixels, maximum 16,777,216 pixels; non-thinking processor \\
Parallelism and hardware & 8 nodes with 8 GPUs each, 64 GPUs total; tensor parallelism 2; GPU model not recovered \\
\bottomrule
\end{tabularx}
\caption{Actual configuration associated with the step-473 training run, cross-checked against command overrides, optimizer logs, and checkpoint records.}
\label{tab:sft-configuration}
\end{table*}

The nominal number of data passes follows from $473\times256/60{,}517\approx2.001$. It describes the configured budget and does not replace per-sample observed exposure counts. SHA-256 digests are preserved separately for the actual input file, merged trajectory manifest, launch configuration, and evaluation configuration; model usage and effective training tokens require separate log statistics.

\paragraph{Provenance and overlap checks.}
All training data come from \system{}'s historical interaction-collection batches. Exact matching after lowercasing and word normalization across 5,856 training instructions, 102 REAL tasks, and 200 Odysseys tasks found no identical text. For test instructions with at least eight words, screening for word-set Jaccard similarity of at least 0.65 also found no candidates (100 REAL tasks and 200 Odysseys tasks). These checks do not cover semantic paraphrases, shared task templates, or test-derived relationships.

Of the 74 training site labels, 69 map to source hosts through the current resource table. Comparing evaluation entry points against literal hostnames after removing the \texttt{www.} prefix found no matching host overlap for REAL. The intersection of all Odysseys entry points with training sources includes Amazon, Apple, Google, and YouTube. This check does not cover every host visited during task execution, nor does it map REAL application aliases to unconfirmed original-site domains. Source-host overlap does not imply use of the same mirror version, and different entry points do not guarantee cross-site isolation. This comparison uses the step-473 checkpoint at the training endpoint.

\paragraph{REAL.}
The full set contains 102 tasks, with 11 infrastructure failures for the post-trained model. Main results use the 91 tasks with valid evaluation results for both models: the baseline succeeds on 22/91 (24.2\%) and the post-trained model on 31/91 (34.1\%), a net gain of nine tasks (+9.9pp). Per-task paired outcomes comprise 15 failure$\rightarrow$success transitions, six success$\rightarrow$failure transitions, and 70 unchanged outcomes, totaling 91. The complete-denominator robustness analysis retains all 102 tasks and does not count the 11 infrastructure failures as successes: 22/102 (21.6\%)$\rightarrow$31/102 (30.4\%), +8.8pp. The two denominators are reported separately, and per-task changes use descriptive counts. Both models use matching task-selection digests.

\paragraph{Odysseys-Easy.}
Both models use the same harness in the hybrid setting described above. The baseline completes 45/45 tasks; the post-trained model has 44 valid results and one infrastructure failure. The primary comparison uses the 44 tasks with valid results for both models. Baseline and post-trained Rubric Avg are 19.71\% and 25.96\% (+6.25pp); Task Avg is 18.86\% and 25.18\% (+6.32pp); Perfect is 2/44 (4.55\%) and 4/44 (9.09\%). The Perfect gain is computed from raw counts: $100(4/44-2/44)$ is approximately +4.55pp.

The baseline's full 45-task reference is reported separately as Rubric Avg 21.13\%, Task Avg 20.22\%, and Perfect 2/45 (4.44\%); it does not replace the baseline on the 44 mutually valid tasks. The post-trained model's 45th task remains an infrastructure failure. The 44 tasks contain 208 rubric conditions. Rubric Avg is a condition-level micro-average, with baseline and post-trained counts of 41/208 and 54/208. Task Avg is the macro-average of per-task rubric means, and Perfect requires every condition of a task to be satisfied. Aggregates are recomputed from each model's saved results without mixing records from other judges. Results cover the Easy subset of Odysseys.

\paragraph{WebArena-Verified.}
We evaluate the same trained checkpoint used for REAL and Odysseys-Easy with our own implementation of the fixed harness. Within WebArena-Verified, the baseline and SFT models use matched task sets, execution budgets, and evaluation settings. Table~\ref{tab:post-training-utility} retains the benchmark's strict scores. Supplementary rescoring of the same SFT outputs ignores JSON-format judgment: Retrieve passes increase from 39/325 (12.00\%) to 105/325 (32.31\%), a difference of 66 passes (+20.31pp). This is a scoring-protocol comparison without new agent execution. It does not replace the strict comparison or establish the cause of the entire strict Retrieve decline; no format-relaxed baseline or Overall result is reported.

\paragraph{Experimental reference and task distribution.}
The baseline without additional training provides a model reference from the same starting point; this comparison measures transfer utility after post-training on environment data. Benchmarks are reported independently. Training data are generated through a unified task--evaluator pipeline, with limited coverage of long-horizon, multistage tasks and tasks with strong intermediate-state dependencies. This limitation concerns data-construction capabilities; source fidelity is measured separately in RQ2.

\subsection{Data-Source Comparison on Odysseys-All}
\label{sec:data-source-comparison-protocol}

This experiment compares three separately fine-tuned models using the same backbone, training configuration, and evaluation harness. The \system{}, WebHarbor, and VeriEnv collections are identified as \texttt{all661}, \texttt{Core16}, and \texttt{VeriEnv}, respectively. They are separate from the 5,856-trajectory training collection reported in Appendix~\ref{sec:transfer-protocol}. The collections have similar trajectory counts, with their different training-row counts reported in Table~\ref{tab:data-source-training-size}.

\paragraph{Data construction.}
\textbf{WebHarbor (Core16)} starts from 663 native queries across WebHarbor's official 16 sites. Existing trajectories collected with the fixed harness are filtered for successful execution, excluding timeouts and tool errors, leaving 617 trajectories and 5,240 training rows.
\textbf{\system{} (all661)} uses queries generated by our pipeline, matching the site and task-category quotas of the official 663-query WebHarbor pool. Its final training collection contains 661 trajectories and 5,520 rows.
\textbf{VeriEnv} starts from 2,500 native tasks across Apartments, Coursera, Discogs, LinkedIn, and Weather. Task-category quotas select 663 tasks, yielding 663 training trajectories and 9,244 rows. Site-quota matching applies to the \system{}--WebHarbor construction; VeriEnv uses its own five-site task pool.

\begin{table}[!htbp]
\centering
\begingroup
\small
\setlength{\tabcolsep}{4pt}
\renewcommand{\arraystretch}{1.12}
\begin{tabularx}{\linewidth}{Xrr}
\toprule
Training data & Trajectories & Training rows \\
\midrule
WebHarbor (Core16) & 617 & 5,240 \\
\system{} (all661) & 661 & 5,520 \\
VeriEnv & 663 & 9,244 \\
\bottomrule
\end{tabularx}
\caption{Final training-set sizes for the data-source comparison. Row counts refer to training-file records.}
\label{tab:data-source-training-size}
\endgroup
\end{table}

\paragraph{Evaluation population and metrics.}
The evaluation uses Odysseys-All rather than the Easy subset. The reported validity filter excludes infrastructure failures, source-blocked runs, and unfinished runs, leaving 194 valid tasks for \system{}, 196 for WebHarbor, and 197 for VeriEnv. The intersection contains 189 tasks and defines the common comparison in Table~\ref{tab:data-source-comparison}. Table~\ref{tab:data-source-individual-valid} additionally reports results on each model's own valid-task set.

On the common subset, \system{}'s Rubric Avg exceeds WebHarbor and VeriEnv by 3.51 and 4.02 percentage points; its Task Avg is higher by 3.91 and 4.17 percentage points, respectively. Under shared training and evaluation settings, the \system{}-trained model achieves higher Rubric Avg and Task Avg than models trained on the compared collections.

Rubric Avg pools satisfied rubric conditions across the retained tasks. Task Avg equally weights each retained task's rubric-completion fraction. Perfect requires every rubric condition of a task to be satisfied. All reported scores use the retained valid-task sets from the 200-task suite.

\begin{table}[!htbp]
\centering
\begingroup
\small
\setlength{\tabcolsep}{3pt}
\renewcommand{\arraystretch}{1.12}
\begin{tabularx}{\linewidth}{Xrrr}
\toprule
Metric & \system{} & WebHarbor & VeriEnv \\
\midrule
Valid tasks & 194 & 196 & 197 \\
Rubric Avg & 10.73 & 6.98 & 6.62 \\
Task Avg & 12.12 & 7.81 & 7.77 \\
Perfect & 0.52 & 0.00 & 0.00 \\
Perfect count & 1/194 & 0/196 & 0/197 \\
\bottomrule
\end{tabularx}
\caption{Odysseys-All results using each model's own valid-task set. Score values are percentages. The common 189-task comparison is reported separately in Table~\ref{tab:data-source-comparison}.}
\label{tab:data-source-individual-valid}
\endgroup
\end{table}

\section{Reproducibility Materials and Data Governance}
\label{sec:environment-atlas}

Supplementary materials include the 244-site inventory, the 14-domain/53-category mapping, additional-review markers for 113 sites, and original $1440\times1000$ screenshots and digests. The separate atlas retains all frozen homepages by application domain without filtering presentation or resource anomalies. Historical qualitative comparisons additionally preserve nine original screenshots and their corresponding per-page scores.

For reproduction, the resource pool, historical 85-site subset, RQ2 comparison artifacts, and training sources are associated separately. Models, harnesses, tasks and evaluators, runtime budgets, initial states, test adapters, and reasons for nonexecution are preserved. The evidence scope of each result appears in Appendix~\ref{sec:result-provenance}.

Environment distribution is constrained by content licenses, privacy, and access authorization; restricted resources are handled separately from publicly shareable anonymized statistics. Side-effecting operations involving payments, posting, deletion, or real-account modification are prohibited by default. Released environments contain neither real credentials nor private user content. AI tools assisted textual organization and formatting of this manuscript; the authors are responsible for verifying experimental numbers and technical conclusions.

\subsection{Complete Site Inventory}
\label{sec:environment-site-list}

The following lists all 244 unique site artifacts frozen as of October 9, 2026, grouped by first-level application domain in the resource inventory. Parentheses indicate site counts per domain; there are 14 domains and 53 fine-grained categories. Names and categories follow the frozen inventory. The list describes resource-pool inclusion, not uniform functional acceptance of every site. Artifacts with presentation anomalies are retained; diagnostic definitions appear in Appendix~\ref{sec:pool-quality-depth}.

\begingroup
\small
\raggedright

\paragraph{E-commerce and retail (33 sites).}
Amazon; Apple; Backcountry; Best Buy; B\&H Photo; Birkenstock; CarMax; Cars.com; Carvana; Dillard's; DJI; Google Shopping; IKEA; JCPenney; Kelley Blue Book; Macy's Wine Shop; Michaels; Micro Center; Parkers; Porsche; Prestashop; Samsung; SHEIN; Shopify; Speedo; Target; Trader Joe's; UNIQLO; US Appliance; Versus; Warhammer; Wine Access; Zara.

\paragraph{Development and research (32 sites).}
Anthropic; ArXiv; Choosealicense; Claude Code; Cohere; DBLP; DeepSeek; DJICDN; ElevenLabs; Gitea; GitHub; Gitlab; Google DeepMind; Grafana; Groq; Home Assistant; Hugging Face; ICLR; CERN WWW Project; Kaggle; Microsoft Azure; Mistral; ModelScope; Tencent Cloud; NVIDIA; Openreview; Phys.org; SourceForge; Statista; Superset; Wandb; Zotero.

\paragraph{Travel and local services (30 sites).}
Airbnb; Amtrak; Apartments.com; Booking; Carnival Cruise; China Eastern Airlines; Compass; China Southern Airlines; Ctrip; Delta; Doordash; FlightAware; Google Flights; LandWatch; Marriott; Megabus; MTA; Opentable; Public Storage; Qatar Airways; Raising Cane's; RE/MAX; Ryanair; SpotHero; Student.com; Thumbtack; TourRadar; Trip.com; United Airlines; Wanderlog.

\paragraph{Finance and enterprise services (26 sites).}
ACCA; Amazon Jobs; American Express; Better Business Bureau; Cboe; Chase; Chronicle Jobs; Chinese Institute of Certified Public Accountants; CoinMarketCap; Erpnext; Google Finance; Hubspot; IRS Refund; NYSE; OhioMeansJobs; Robinhood; U.S. Securities and Exchange Commission; Shui5; SmartAsset; Super Lawyers; Turbotax; Walmart Careers; Y Combinator; Yahoo Finance; Zendesk; ZipRecruiter.

\paragraph{Entertainment, sports, and media (26 sites).}
Bandcamp; BBC News; Bilibili; BoardGameGeek; Chess.com; Discogs; Disney; ESPN; Eventbrite; Fox Sports; IGN; IMDb; Jellyfin; League of Legends; NBA; NFL; Playblackdesert; Red Bull; Rotten Tomatoes; SoundCloud; Stashmedia; Steam; StubHub; Ticketek; Ticketmaster; YouTube.

\paragraph{Government and public services (17 sites).}
ACRA Singapore; National Audit Office of the People's Republic of China; California.gov; Canada Revenue Agency; Data.gov; GOV.UK; Library of Congress; Medicare.gov; NPS.gov Homepage (U.S. National Park Service); Ohio.gov; Recreation.gov; U.S. Customs and Border Protection; U.S. Department of Justice; USA.gov; USCIS; United States DOGE Service; Virginia DMV.

\paragraph{Search, knowledge, and general-purpose tools (15 sites).}
AccuWeather; Cambridge Dictionary; Drupal; Ghost; Google Maps; Google Search; Huiji Wiki; Merriam-Webster; Tampermonkey; The Weather Network; Weather.com; Wikipedia; Windy; Wolfram Alpha; Wordpress.

\paragraph{Collaboration and productivity (13 sites).}
Airtable; Asana; Canva; Dropbox; Excalidraw; Figma; Linear; Nextcloud; Notion; Penpot; Surveymonkey; Trello; WPS.

\paragraph{Education (13 sites).}
UC Berkeley; Civil Aviation Flight University of China; Civil Aviation University of China CUCAS; Coolmath4Kids; Coursera; Instructure; Moodle; Ohio State University; PhET Simulations; Shanghai Innovation Institute; Stanford University; TED; University of Michigan.

\newpage
\paragraph{Communication and office tools (11 sites).}
4shared; Discord; Feishu; Gmail; Google Calendar; Jitsi; Mattermost; MEGA; Rocket Chat; Slack; Verizon.

\paragraph{Social networking and communities (11 sites).}
9GAG; Craigslist; Discourse; Fandom; Imgur; LinkedIn; Mastodon; Reddit; Threads; Tumblr; X (Twitter).

\paragraph{Healthcare (9 sites).}
America's Health Rankings; BabyCenter; CVS; Drugs.com; Healthgrades; Healthline; Mayo Clinic; WebMD; WebMD Doctor.

\paragraph{Lifestyle Knowledge (5 sites).}
Adopt-a-Pet; American Kennel Club; Allrecipes; Cookpad; Petfinder.

\paragraph{Logistics services (3 sites).}
FedEx; UPS; USPS.

\endgroup

\end{document}